\documentclass{article}

\PassOptionsToPackage{numbers, compress}{natbib}
\usepackage[preprint]{neurips_2024}

\usepackage[utf8]{inputenc} % allow utf-8 input
\usepackage[T1]{fontenc}    % use 8-bit T1 fonts
\usepackage{hyperref}       % hyperlinks
\usepackage{url}            % simple URL typesetting
\usepackage{booktabs}       % professional-quality tables
\usepackage{amsfonts}       % blackboard math symbols
\usepackage{nicefrac}       % compact symbols for 1/2, etc.
\usepackage{microtype}      % microtypography
\usepackage{xcolor}         % colors

\usepackage{booktabs}  % Required for \toprule, \midrule
\usepackage{graphicx}  % Required for \resizebox
\usepackage{multirow}  % Required for \multirow

\usepackage{amsmath}
\usepackage{amssymb}
\usepackage{mathtools}
\usepackage{amsthm}

\usepackage{wrapfig}
\usepackage{algorithm}
\usepackage{algpseudocode}
\usepackage{subcaption}
\usepackage{enumitem}
\usepackage{multicol, multirow}
\usepackage{threeparttable}

\title{Toward a Better Understanding of Fourier Neural Operators from a Spectral Perspective}

\author{
    Shaoxiang Qin$^{1 2}$\thanks{Equal contribution} \thanks{Work done as a research assistant at Concordia University} , Fuyuan Lyu$^{2}$\footnotemark[1], Wenhui Peng$^{3}$, Dingyang Geng$^{1}$, Ju Wang$^{4}$, Xing Tang$^{5}$,\\ 
    \textbf{Sylvie Leroyer}$^{6}$\textbf{, Naiping Gao}$^{7}$\textbf{, Xue Liu$^{2}$}\textbf{, Liangzhu (Leon) Wang}$^{1}$ \\
    $^{1}$Concordia University, $^{2}$McGill University,\\ $^{3}$The Hong Kong Polytechnic University, 
    $^{4}$Northwest University Xi'an,\\ 
    $^{5}$FiT, Tencent, $^{6}$Environment and Climate Change Canada, $^{7}$Tongji University \\
    \texttt{leon.wang@concordia.ca} \\
}

\begin{document}

\maketitle
\begin{abstract}
In solving partial differential equations (PDEs), Fourier Neural Operators (FNOs) have exhibited notable effectiveness.
However, FNO is observed to be ineffective with large Fourier kernels that parameterize more frequencies.
Current solutions rely on setting small kernels, restricting FNO's ability to capture complex PDE data in real-world applications.
This paper offers empirical insights into FNO's difficulty with large kernels through spectral analysis: FNO exhibits a unique Fourier parameterization bias, excelling at learning dominant frequencies in target data while struggling with non-dominant frequencies. 
To mitigate such a bias, we propose SpecB-FNO to enhance the capture of non-dominant frequencies by adopting additional residual modules to learn from the previous ones' prediction residuals iteratively. 
By effectively utilizing large Fourier kernels, SpecB-FNO achieves better prediction accuracy on diverse PDE applications, with an average improvement of 50\%.
\end{abstract}
\section{Introduction}
\label{sec:intro}

In natural sciences, partial differential equations (PDEs) serve as fundamental mathematical tools for modeling and understanding a wide range of phenomena, such as fluid dynamics~\cite{temam2001navier}, heat conduction~\cite{kittel1998thermal,FourCastNet}, and quantum mechanics~\cite{messiah2014quantum}. 
Traditionally, numerical simulations of PDEs are employed to analyze complex physical processes.
However, solving PDEs with numerical simulators requires substantial time and computational cost, given their fine granularity. 

% Machine learning offers promising alternatives for numerical solvers by proposing more efficient surrogate models~\cite{deeponet,FNO,NO-JMLR,MP-PDE}. 
% In particular, Fourier Neural Operator (FNO)~\cite{FNO} emerged among surrogate models due to its superior precision~\cite{FNO}, universal approximity~\cite{PsiFNO}, and resolution-invariant property~\cite{FNO}.
Machine learning offers promising alternatives for numerical solvers by proposing more efficient surrogate models~\cite{deeponet,FNO,NO-JMLR,MP-PDE}. 
In particular, Fourier Neural Operator (FNO)~\cite{FNO} has been applied to solve various realistic PDE problems due to its superior accuracy and resolution-invariant property~\cite{pathak2022fourcastnet, rashid2022learning, yang2021seismic}.
FNO parameterizes its convolution kernels in Fourier space, showcasing notably superior performance compared to traditional convolution-based (Conv-based) networks~\cite{resnet,U-Net}, which parameterize their convolution kernels in spatial space. 
% Notable improvements have also been made upon FNO for reducing model complexity~\cite{FFNO}, irregular grid~\cite{G-FNO,GINO}, and better generlizability~\cite{CAPE}. However, the performance gap between FNOs and numerical solvers remains large~\cite{FFNO}, limiting its application in the real world.

% However, FNO still suffers from several limitations. On the one hand, FNO, like other neural networks, suffers from spectral bias~\cite{HANO}. On the other hand, its incremental modes can not significantly increase, limiting its capability to capture high-frequency details.

Despite significant improvements in FNO's accuracy across various scenarios, one challenge remains unsolved: \textit{FNO is ineffective with larger Fourier kernels that cover a wider range of frequencies.} 
% These frequencies usually contain detailed local features, which proves to be important in natural science. 
The current solution involves setting relatively small Fourier kernels manually~\cite{FNO, FFNO, G-FNO, DAFNO} or automatically~\cite{iFNO}, thereby restricting FNO's ability to capture more complex PDE data in the real world.

% In this paper, we aim to propose a more effective surrogate model. To obtain meaningful intuitions on this issue, we take a step back and investigate why FNOs emerge among surrogate models. Specifically, we conduct a spectral analysis over FNO in Figure \ref{fig:intro_1}. The energy distribution of PDE data is more concentrated in Fourier space than in spatial space, as shown in Figure \ref{fig:intro_1} (a) and (b). 
% In this research, we first present empirical evidence from a spectral perspective regarding the reasons behind FNO's superiority over Conv-based networks: FNO's capacity to capture dominating frequencies in the target data significantly surpasses Conv-based surrogate models. 

To address this issue, we need a deeper understanding of why FNO cannot benefit from larger Fourier kernels. In this paper, we conduct a spectral analysis of FNO and first identify the spectral property of FNO that explains its drawback when employing large kernels: FNO struggles to learn target data's non-dominant frequencies within its Fourier kernel effectively.
We summarize FNO's unique spectral property as follows:

\textbf{Fourier parameterization bias.} \textit{Compared to convolution kernels parameterized in spatial space, convolution kernels parameterized in Fourier space exhibit a stronger bias toward the dominating frequencies in the target data.}

\begin{wrapfigure}{o}{0.6\textwidth}
  \centering
  \includegraphics[width=0.6\textwidth]{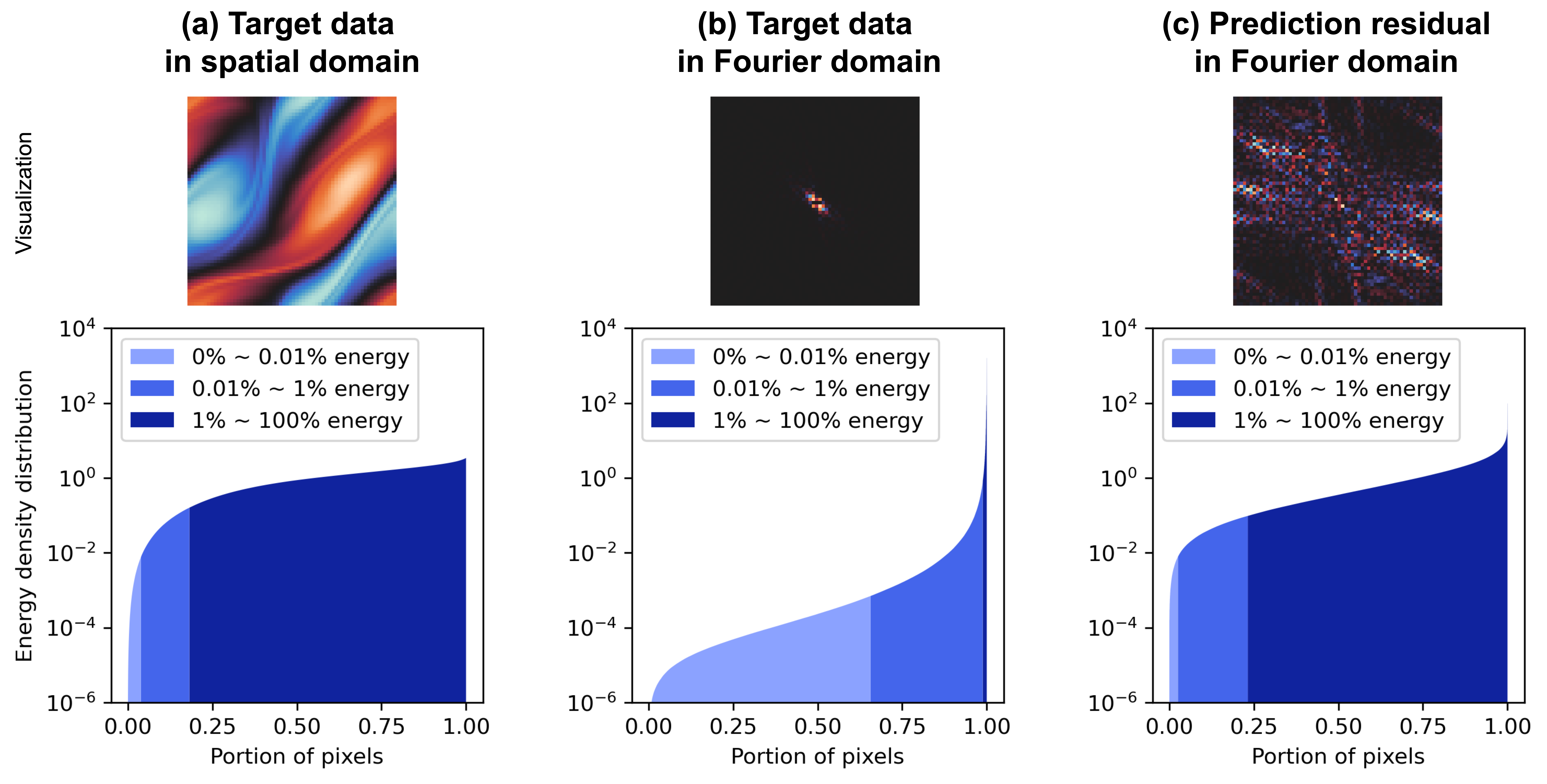}
  % \vspace{-10pt}
  \caption{Energy density distribution of pixels with small to large features on Navier-Stokes ($\nu$ = 1e-5). In Fourier space, the energy distribution of the target data is more concentrated than in spatial space. Specifically, 1.2\% of the pixels with larger features contain 99\% of the energy in Fourier space. In contrast, the prediction residual has a more even energy distribution in Fourier space.}
  %\vspace{-15pt}
  \label{fig:intro_1}
\end{wrapfigure}

Fourier parameterization bias is caused by the energy distribution of PDE data being more concentrated in Fourier space than in spatial space, as shown in Figure \ref{fig:intro_1}a and \ref{fig:intro_1}b. As a result, the loss function focuses on optimizing the few dominant frequencies and overlooks the remaining non-dominant frequencies.

To address FNO's Fourier parameterization bias and enhance FNO's ability with a larger Fourier kernel, we introduce \textbf{Spec}tral \textbf{B}oosted \textbf{FNO}, abbreviated as SpecB-FNO, designed to enhance the capture of non-dominant frequencies using multiple neural operators. In SpecB-FNO, following the regular training of an FNO, additional residual modules are trained to predict the residuals of the previous ones. 
The intuition of SpecB-FNO is that the energy of FNO's prediction residuals is more evenly distributed in Fourier space than that of the target data, as shown in Figure \ref{fig:intro_1}b and \ref{fig:intro_1}c.
%Since a single FNO captures dominant frequencies well, it leaves relatively smaller residuals for these frequencies. Conversely, non-dominant frequencies are less well captured, resulting in larger residuals. As a result, the energy of the FNO's prediction residuals is more evenly distributed in Fourier space than that of the target data, as shown in Figure \ref{fig:intro_1} (b) and (c). Thus, SpecB-FNO mitigates the effect of FNO's Fourier parameterization bias.
% SpecB-FNO is empirically evaluated on five PDE datasets with different characteristics. An up to 72\% reduction in error is witnessed. 
SpecB-FNO is empirically evaluated on five PDE datasets with different characteristics. A reduction of up to 93\% in prediction error is witnessed. SpecB-FNO enables FNO to learn from PDE data with significantly larger Fourier kernels. Additionally, SpecB-FNO proves to be a memory-efficient solution for training larger surrogate FNOs.
% Additionally, SpecB-FNO proves to be a memory-efficient solution for training larger surrogate FNOs and a feasible solution for effectively capturing high-frequency data. 

% To overcome the downside effect of FNO's Fourier parameterization bias and further increase the capability of FNO, we introduce \textbf{m}ulti-stage \textbf{r}esidual-learning FNO (MR-FNO), a framework designed to enhance the capture of non-dominant frequencies by adopting additional FNOs iteratively learning on the residual of the previous one. Since FNO's Fourier parameterization bias ensures its capability to capture dominant frequencies well, it leaves relatively smaller residuals for the dominant frequencies. Conversely, non-dominant frequencies are less well captured, resulting in larger residuals. As a result, the energy of the FNO's prediction residuals is more evenly distributed in Fourier space than that of the target data, as shown in Figure \ref{fig:intro_1} (b) and (c). Therefore, MR-FNO mitigates the downside of FNO's Fourier parameterization bias. MR-FNO is empirically evaluated on five PDE datasets with different characteristics. An up to 72\% reduction in error is witnessed. Additionally, MR-FNO proves to be a memory-efficient solution for training larger surrogate models and a feasible solution for effectively capturing high-frequency data. 
Our contributions can be summarized as follows:
\begin{itemize}[topsep=0pt,noitemsep,nolistsep,leftmargin=*]
    \item By utilizing spectral analysis on the model prediction error, we identify the Fourier parameterization bias of FNO, which empirically explains FNO's incompatibility with large Fourier kernels. 
    \item To address FNO's Fourier parameterization bias, we propose SpecB-FNO, which enables training FNO with large Fourier kernels.
    \item We validate SpecB-FNO's superiority on various PDE applications. Compared to the best-performing baselines, SpecB-FNO achieves an average error reduction of 50\%.
\end{itemize}

\section{Spectral Properties of Fourier Neural Operator}
\label{sec:spectral}

\begin{figure}
\vspace{-10pt}
\centering
\begin{subfigure}{0.32\textwidth}
    \includegraphics[width=\textwidth]{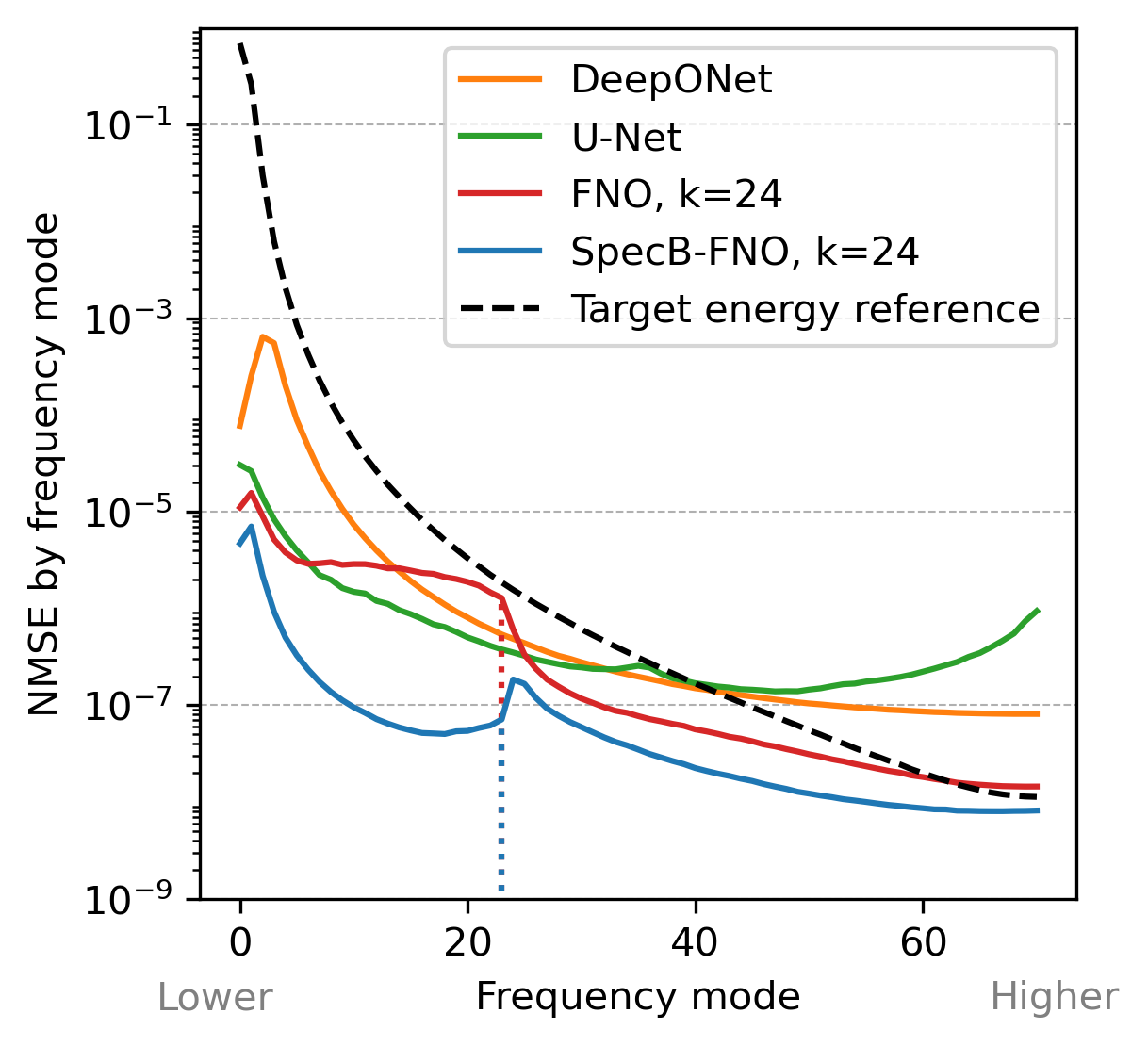}
    \caption{Darcy flow}
    \label{fig:spectral-darcy}
\end{subfigure}
\hfill
\begin{subfigure}{0.32\textwidth}
    \includegraphics[width=\textwidth]{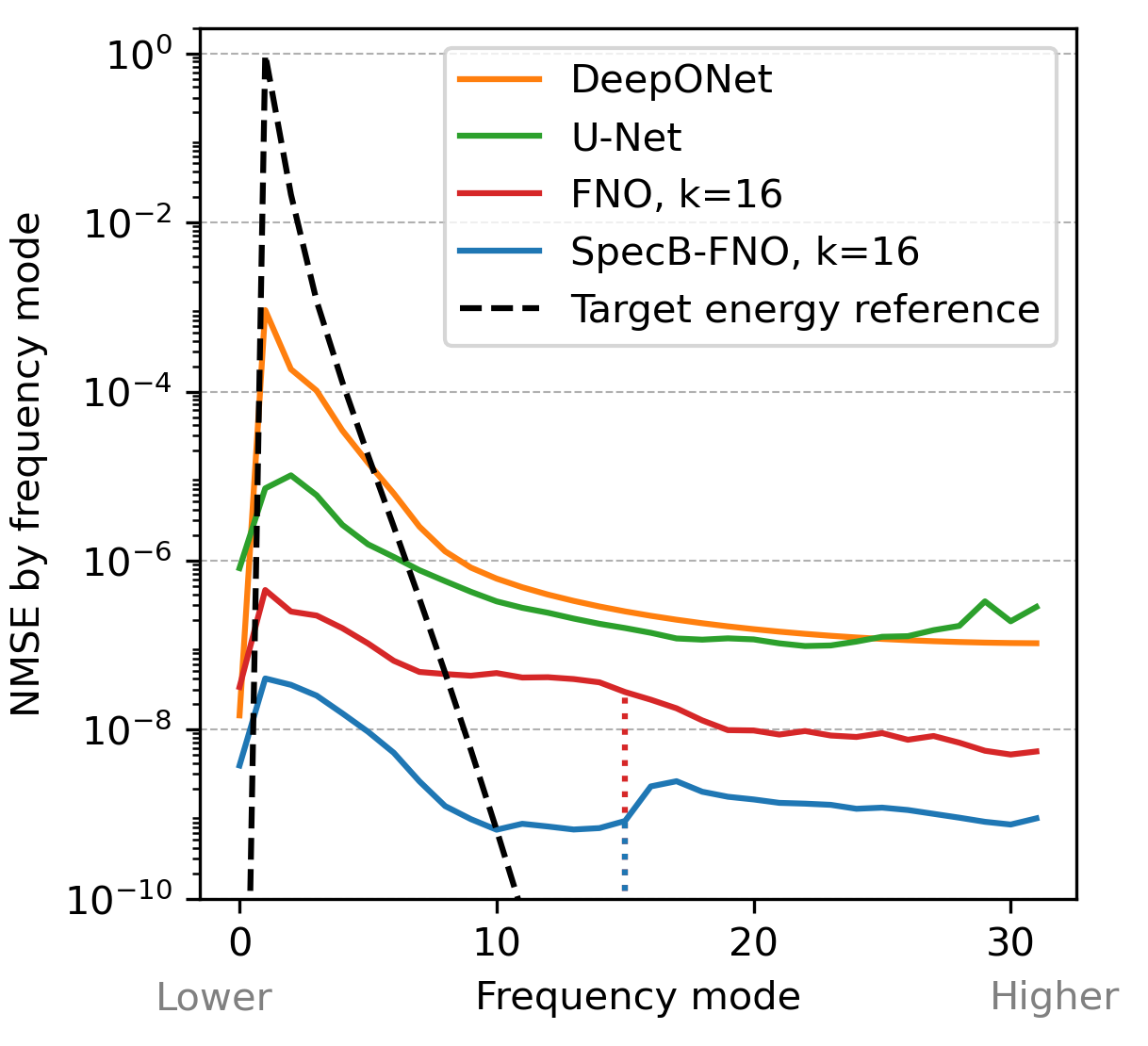}
    \caption{Navier-Stokes ($\nu$ = 1e-3)}
    \label{fig:spectral-ns-3}
\end{subfigure}
\hfill
\begin{subfigure}{0.32\textwidth}
    \includegraphics[width=\textwidth]{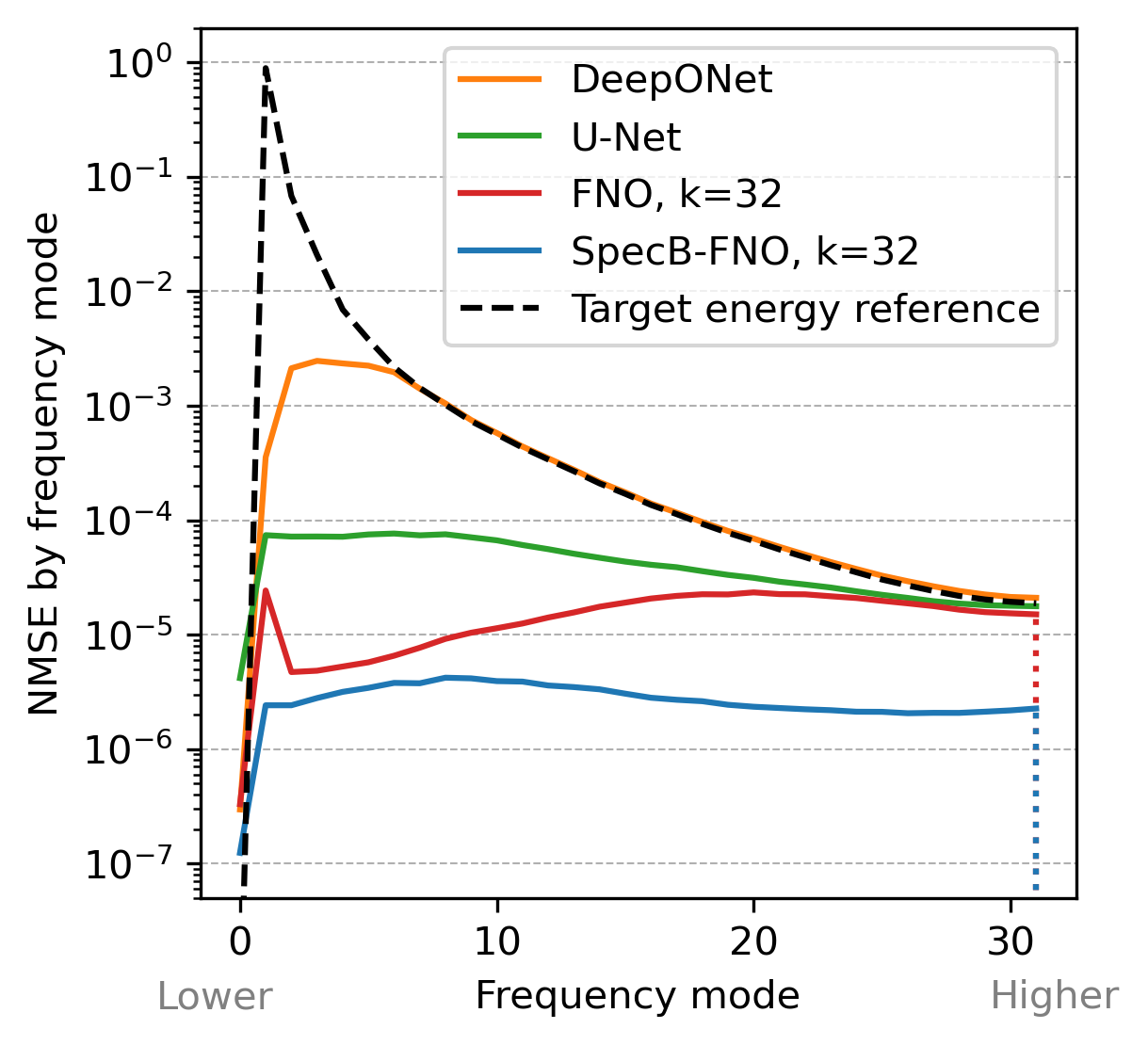}
    \caption{Navier-Stokes ($\nu$ = 1e-5)}
    \label{fig:spectral-ns-5}
\end{subfigure}
\hfill
\begin{subfigure}{0.32\textwidth}
    \includegraphics[width=\textwidth]{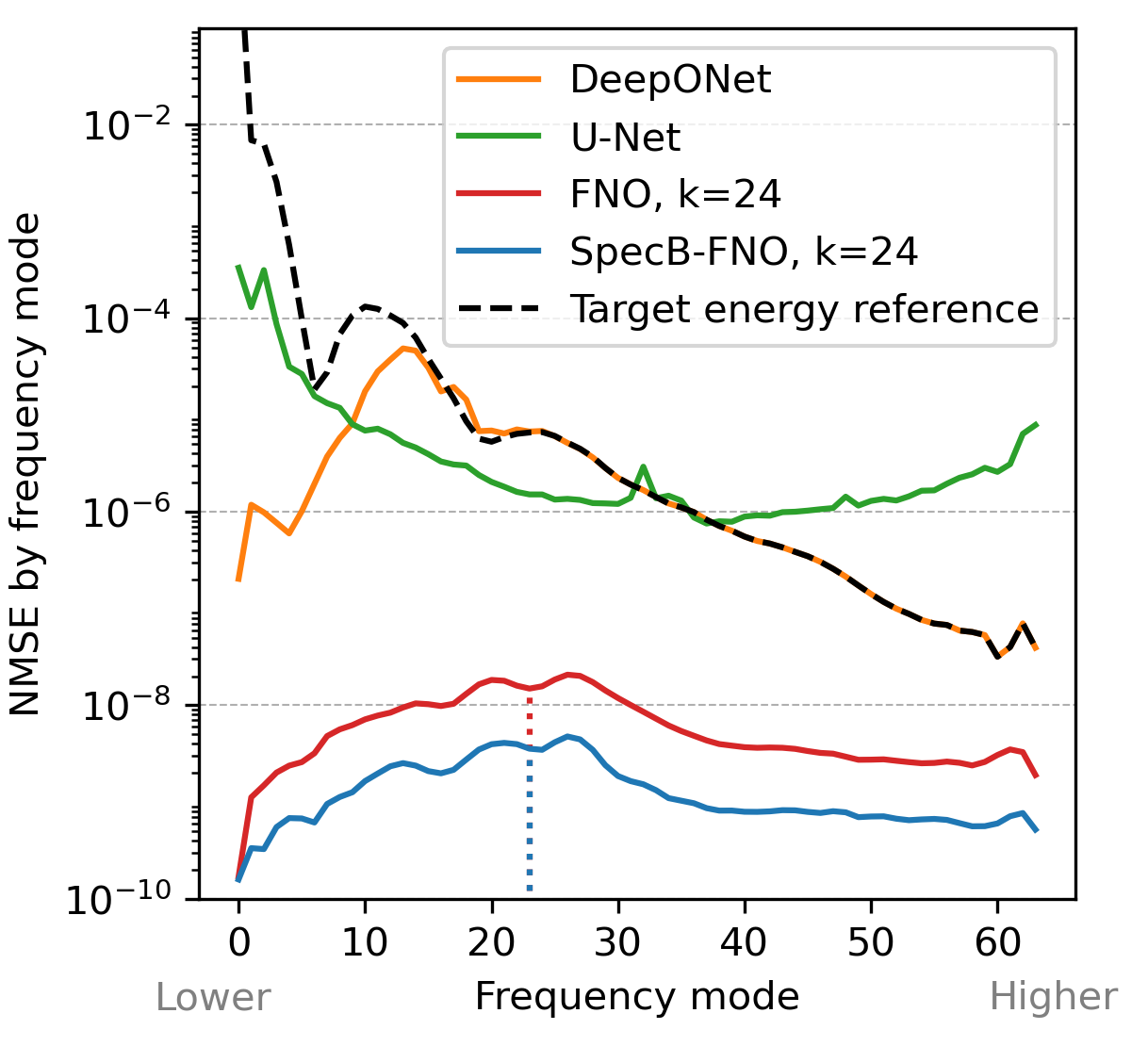}
    \caption{Shallow water}
    \label{fig:spectral-shallow}
\end{subfigure}
\hfill
\begin{subfigure}{0.32\textwidth}
    \includegraphics[width=\textwidth]{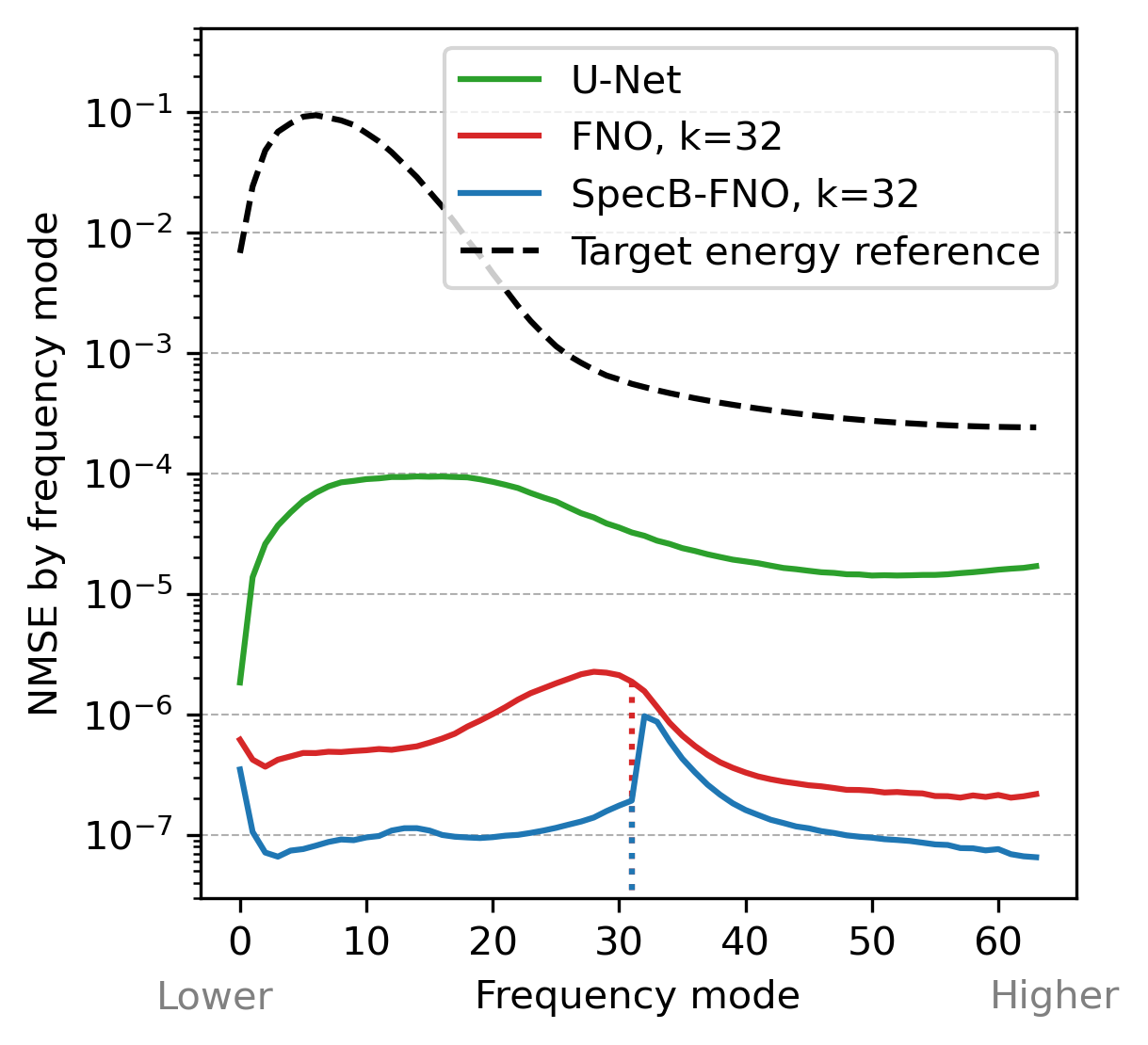}
    \caption{Diffusion-reaction (activator)}
    \label{fig:spectral-diffusion-a}
\end{subfigure}
\hfill
\begin{subfigure}{0.32\textwidth}
    \includegraphics[width=\textwidth]{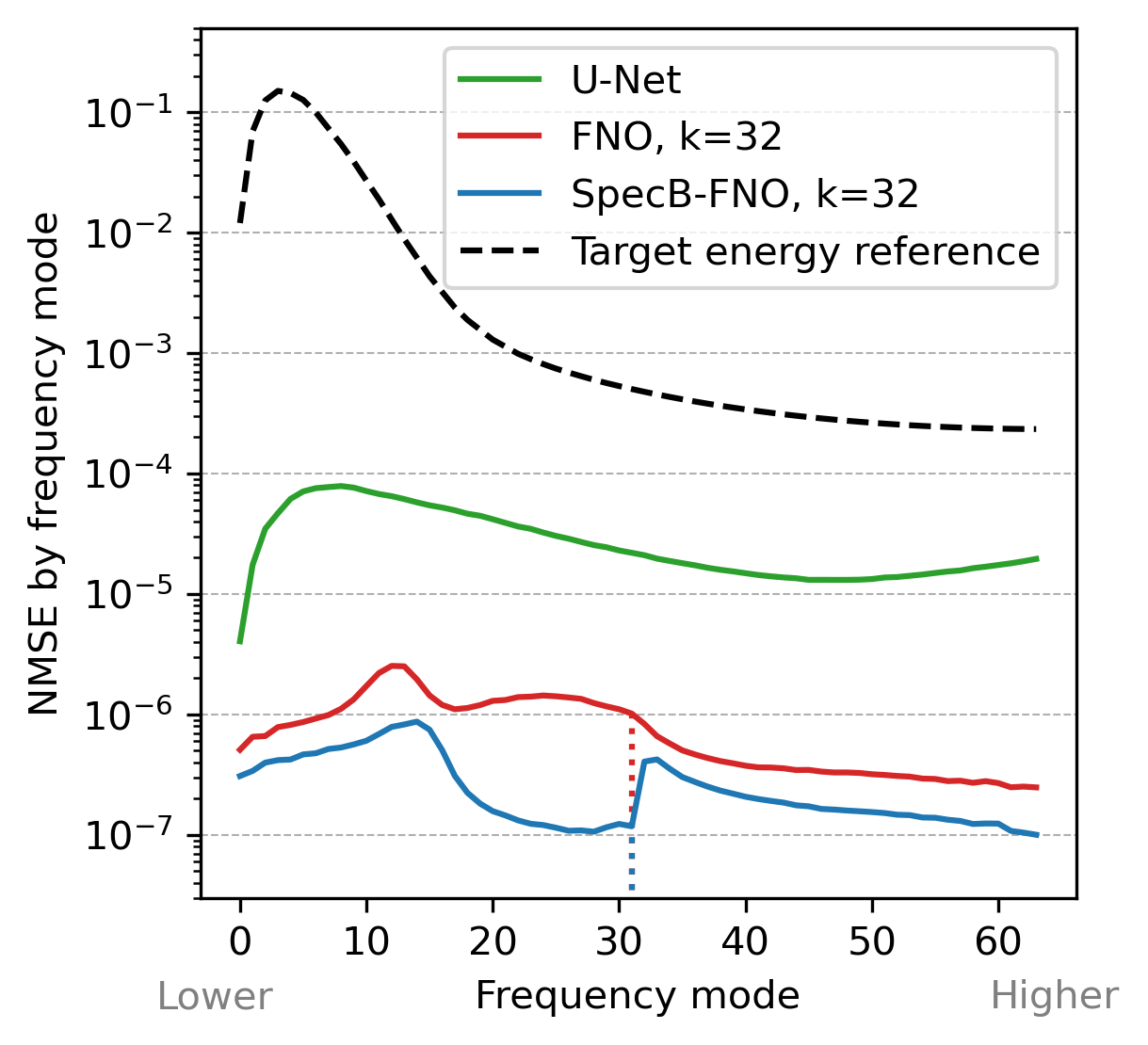}
    \caption{Diffusion-reaction (inhibitor)}
    \label{fig:spectral-diffusion-i}
\end{subfigure}
\caption{NMSE spectrums on different PDE datasets. FNO's truncation frequency, k, is marked with a dotted line. The target energy reference is the energy spectrum of the target data, providing information on dominating frequencies. Two features of the diffusion-reaction equation (activator and inhibitor) are presented separately due to their different dominating frequencies.}
\vspace{-10pt}
\label{fig:spectral}
\end{figure}

% FNO exhibits different spectral performances below and beyond the truncation frequency.

% Within the truncation frequency, FNO demonstrates a unique Fourier parameterization bias. 
In this section, we analyze the spectral properties of surrogate models for learning PDEs to empirically demonstrate FNO's Fourier parameterization bias. We choose DeepONet~\cite{deeponet} as the representative MLP-based model and U-Net~\cite{U-Net} as the representative Conv-based model. They both serve as widely used baselines in literature~\cite{FNO, gupta2022towards, G-FNO, CNO, CORAL}.  

To demonstrate the spectral property of a surrogate model, we decompose its prediction residual (the difference between the target and the model prediction) into Fourier space and show how the energy is distributed across different frequencies. We refer to this curve as the NMSE spectrum, as the sum of the energy spectrum equals the normalized mean squared error (NMSE) of the model prediction:
\begin{equation}
\label{eq:nmse}
\begin{aligned}
    \text{NMSE} = \frac{1}{|\mathcal{D}|}\sum_{(x, y) \in \mathcal{D}} \frac{||\hat{y} - y||_2^2}{||y||_2^2}, \quad
    \hat{y} = \mathcal{G}(x). 
\end{aligned}
\end{equation}
This follows from Parseval's theorem~\cite{Parseval}, which states that the energy of a signal remains conserved during the discrete Fourier transform. 
Detailed calculation of NMSE spectrum is shown in Appendix \ref{sec:nmse_spectrum_computation}.
% {\color{red} Note that we do not use the training loss NRMSE in Eqn.~\eqref{eq:rmse} for the spectrum because the L2-norm of the signal is not preserved during the Fourier transform.}

In Figure \ref{fig:spectral}, we present the NMSE spectrum of predictions from DeepONet, U-Net, and FNO across various PDE datasets. These datasets are commonly used as benchmarks in neural operator research. They encompass a range of important PDEs with different properties (details in Appendix \ref{sec:appendix_dataset}). Each figure also includes a target energy reference, which is the energy spectrum of the target ground truth on the same axis, allowing us to identify the dominant frequencies in the target data. For example, the dominant frequencies for the PDEs in Figures \ref{fig:spectral-darcy} to \ref{fig:spectral-shallow} are near frequency modes 0 and 1, while the dominant frequencies for the diffusion-reaction equation in Figures \ref{fig:spectral-diffusion-a} and \ref{fig:spectral-diffusion-i} are relatively higher, between frequency modes 5 and 10.
Two observations can be drawn from Figure \ref{fig:spectral}.
% We draw two main observations from Figure \ref{fig:spectral}, which provide a better understanding of FNO's spectral performance:

\paragraph{Observation 1: FNO exhibits different spectral performances below and above its truncation frequency.}
% (1) FNO exhibits different spectral performances below and above its truncation frequency.

For each PDE in Figure \ref{fig:spectral}, the truncation frequency mode $k$ of an FNO is marked with a dotted line. It's evident that the NMSE spectrum trend for FNO differs below and above its truncation frequency, while DeepONet and U-Net show more consistent trends across different frequencies. This is due to FNO's design, which truncates higher frequencies and only parameterizes its Fourier kernels for frequencies lower than $k$. Consequently, frequencies higher than the truncated threshold $k$ are learned by the linear and MLP components. Therefore, FNO's NMSE spectrum beyond its truncation frequency $k$ is similar to that of DeepONet, which also uses MLP for PDE prediction.

\paragraph{Observation 2: Below the truncation frequency, FNO shows a unique Fourier parameterization bias. }
% (2) Below the truncation frequency, FNO shows a unique Fourier parameterization bias. 

Based on Observation 1, we mainly focus on FNO's NMSE spectrum below its truncation frequency, which reflects the spectral property of Fourier kernels.
In Figure \ref{fig:spectral}, compared to U-Net, which parameterizes its convolution kernels in spatial space, FNO shows a stronger bias toward the dominant frequencies in the target data. The greatest relative improvements from U-Net to FNO occur around the dominant frequencies of the target data. For instance, for the PDEs in Figures \ref{fig:spectral-darcy} to \ref{fig:spectral-shallow}, with dominant frequencies near modes 0 and 1, the largest improvement from U-Net to FNO is around the low frequencies. Similarly, for the PDEs in Figures \ref{fig:spectral-diffusion-a} and \ref{fig:spectral-diffusion-i}, with dominant frequencies around modes 5 to 10, the largest improvement from U-Net to FNO occurs around frequencies 5 to 10. 

Thus, we can summarize the common property of FNO across all PDEs: \textit{below the truncation frequency, FNO has a greater capability to learn the dominant frequencies in the target data while being less effective at learning the remaining non-dominant frequencies}. We name such unique spectral performance as the Fourier parameterization bias because the underlying reason for this is parameterizing convolution kernels in Fourier space. As shown in Figure \ref{fig:intro_1}, most of the energy in target data is included in a few dominant frequencies in Fourier space. Since the energy in these dominant frequencies is often exponentially higher than in non-dominant frequencies, FNO focuses on optimizing these dominant frequencies to minimize their prediction errors. 

\paragraph{Why large Fourier kernels are ineffective}
After identifying the Fourier parameterization bias, it becomes clear why FNO cannot benefit from larger Fourier kernels. Even with a larger Fourier kernel, FNO still focuses on a few dominant frequencies and cannot effectively learn the additional parameters to approximate non-dominant frequencies. As a result, the poorly learned non-dominant frequencies will produce noise, consistent with observations in existing research~\cite{iFNO}. Figure \ref{fig:spectral-darcy1}, which shows FNO's NMSE spectrum with increasing truncation frequency on the Darcy flow dataset, validates this hypothesis. The prediction residual does not decrease as the Fourier kernel size increases. 
% For each FNO, the worst-performing frequencies are the relatively higher frequencies within the truncation frequency. 
The error curve for each FNO shows an unusual rise near the higher frequencies within the truncation frequency. These frequencies are the least dominant frequencies associated with the Darcy flow dataset within the Fourier kernel. For example, for FNO with $k$ = 16, the rise occurs around modes 8 to 16, and for FNO with $k$ = 32, it occurs around modes 20 to 32.

The Fourier parameterization bias reveals a key performance bottleneck of FNO with larger Fourier kernels: \textbf{learning non-dominant frequencies} in the target data. This insight motivates us to improve FNO's ability to capture non-dominant frequencies in Section \ref{sec:specbfno}.

\section{SpecB-FNO}
\label{sec:specbfno}

In this section, we first formulate the operator learning and Fourier Neural Operator in Section \ref{sec:specbfno-pde} and \ref{sec:specbfno-fno}. Then, we propose the SpecB-FNO in Section \ref{sec:specbfno-specb}, which mitigates the Fourier parameterization bias to capture non-dominant frequencies and improve prediction accuracy.

\subsection{Operator Learning}
\label{sec:specbfno-pde}

For neural operators, solving PDE is commonly achieved by learning the mapping between continuous functions. Operator learning task aims to predict the output function $\mathcal{Y}$ based on the input function $\mathcal{X}$.
%each representing the state of the partial derivative equation at different time stamps. 
To conduct end-to-end training on surrogate models, function pair $(\mathcal{X}, \mathcal{Y})$ are discretized to instance pair $(x, y)$ during the training process. The objective of PDE data prediction is to learn a surrogate model $\mathcal{G}$ between $(x, y)$, denoted as $y \approx \mathcal{G}(x)$.

Given the training dataset $\mathcal{D} = \{(x, y)\}$, the training objective can generally be formulated as minimizing the normalized root mean square error (NRMSE), which is defined as:
\begin{equation}
\label{eq:rmse}
\begin{aligned}
    \text{NRMSE} = \frac{1}{|\mathcal{D}|}\sum_{(x, y) \in \mathcal{D}} \frac{||\hat{y} - y||_2}{||y||_2}, \quad
    \hat{y} = \mathcal{G}(x), 
\end{aligned}
\end{equation}
where $||\cdot||_2$ represents the L2-norm. Hence, the training objective of PDE data prediction can be summarized as follows:
\begin{equation} \label{eq:pde_obj}
    \min_{(x, y) \in \mathcal{D}} \mathcal{L}_{\text{NRMSE}}(y, \mathcal{G}(x)).
\end{equation}

\subsection{Fourier Neural Operator}
\label{sec:specbfno-fno}

Fourier Neural Operator (FNO) parameterizes its convolution kernel in Fourier space to learn a resolution-invariant mapping between its inputs and outputs. It is one of the most effective surrogate models for learning PDEs.
FNO instantizes the surrogate model $\mathcal{G}$ with the sequential steps of lifting the input channel using $\mathcal{P}$, conducting the mapping through $L$ Fourier layers $\{\mathcal{H}_1, \mathcal{H}_2, \dots, \mathcal{H}_L\}$, and then projecting back to the original channel through $\mathcal{Q}$:
\begin{equation}
\label{eq:fno}
\mathcal{G} =  \mathcal{Q} \circ \mathcal{H}_L \circ \cdots \circ \mathcal{H}_2 \circ \mathcal{H}_1 \circ \mathcal{P}.
\end{equation}
$\mathcal{P}$ and $\mathcal{Q}$ are pixel-wise transformations that can be implemented using models like MLP. 
The key architecture of FNO is its Fourier layer $\mathcal{H}$. In FNO~\cite{FNO}, Fourier layer consists of a linear transformation $\phi(\cdot)$, and an integral kernel operator $\mathcal{K}$:
\begin{equation}\label{eq:fno-l}
\mathcal{H}(x) =  \sigma \left(x + \phi(x) + \text{MLP}(\mathcal{K}(x)) \right),
\end{equation}
with $\sigma$ as the nonlinear activation function, and MLP denotes a multiple-layer perceptron. 
The integral kernel operator $\mathcal{K}$ undergoes a sequential process involving four operations: (i) Fast Fourier Transformation (FFT)~\cite{cochran1967fast}, (ii) high-frequency truncation, (iii) spectral linear transformation, and (iv) inverse FFT. 
Note that various versions of FNO are proposed, detailed in Appendix ~\ref{sec:appendix_fno}, while we adopt the latest and most effective implementation.

\subsection{SpecB-FNO}
\label{sec:specbfno-specb}

\begin{wrapfigure}{o}{0.6\textwidth}
  \centering
  \includegraphics[width=0.6\textwidth]{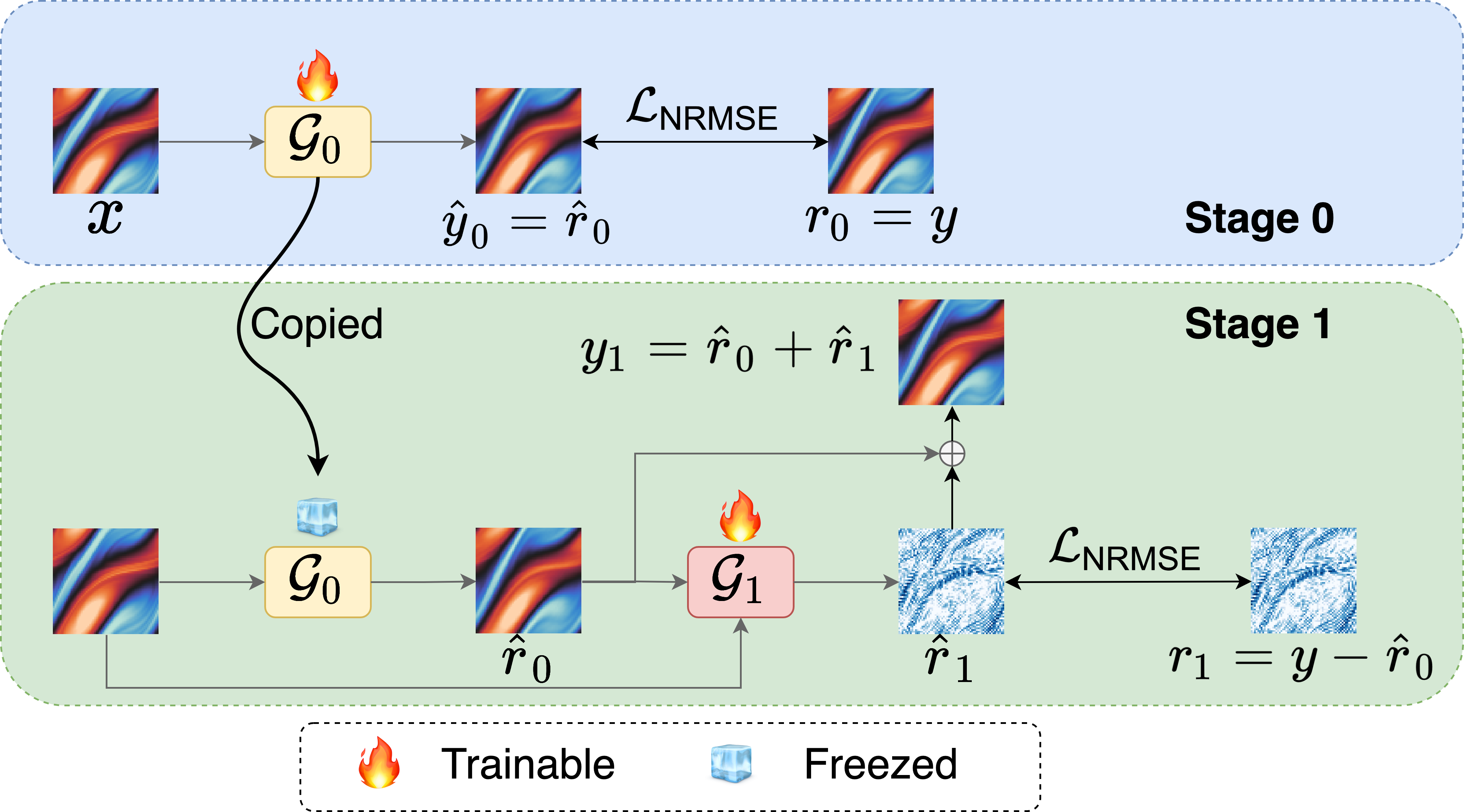}
  % \vspace{-10pt}
  \caption{Illustration of SpecB-FNO with $T=1$.}
  %\vspace{-15pt}
  \label{fig:boost}
\end{wrapfigure}

% \begin{figure}[!htbp]
% \centerline{\includegraphics[width=0.6\columnwidth]{figures/boost2.png}}
% \vspace{-5pt}
% \caption{Illustration of SpecB-FNO with $T=1$.}
% \label{fig:boost}
% \vspace{-5pt}
% \end{figure}

Building upon FNO's Fourier parameterization bias, we propose SpecB-FNO to improve FNO's capability for learning non-dominating frequencies in the target data. SpecB-FNO views each individual FNO as a module and iteratively utilizes an additional module to learn the prediction residual of the previous one. 

The intuition behind SpecB-FNO is that the energy of FNO's prediction residual is more evenly distributed in Fourier space than that of the target data, as shown in Figure \ref{fig:intro_1} (b) and (c). This occurs because a single FNO effectively captures dominant frequencies, leaving relatively smaller residuals for these frequencies. Conversely, non-dominant frequencies are less well captured, resulting in larger residuals. This phenomenon can be observed across all PDE datasets in Figure \ref{fig:spectral}, where we can compare the energy distribution of the target data with that of FNO's prediction residual. In each case, the residual energy distribution is more evenly distributed. Thus, iteratively training additional FNO can effectively mitigate the Fourier parameterization bias. 

After obtaining the initial FNO $\mathcal{G}_0$, which is trained following Eq. \ref{eq:pde_obj}, SpecB-FNO additionally contains $T$ residual modules, which are iteratively trained during $T$ stages. In this paper, we instantize these residual modules as FNO modules with equal configuration as the first FNO module. Without the loss of generalizability, we focus on the $i$-th module(stage) while the rest can be easily generalized. When $T=0$, SpecB-FNO collapses to a naive FNO model in Section \ref{sec:specbfno-fno}.

When training the $i$-th residual module $\mathcal{G}_i(\cdot)$, for each training instance $(x, y) \in \mathcal{D}$, we first calculate the ground truth of the residual for the $i$-th stage $r_{i}$ as follows:
\begin{equation} \label{eq:residual_label}
    r_{i} = y - \sum \hat{r}_{j}, \ 0 \le j \le i-1, \\
\end{equation}
% \begin{equation} \label{eq:residual_label}
% \begin{aligned}
%     r_{i} &= y - \sum y_{(i)} \\
%     y_{(i)} &= [y_0, y_1, \cdots, y_{i}], \ y_0 = \mathbf{0}, \ y_i = \hat{r}_{i-1} \\
% \end{aligned}
% \end{equation}
where $\hat{r}_{j}$ denotes the output of module $j$. For instance, $\hat{r}_0 = \mathcal{G}_0(x)$.
During the $i$-th stage, SpecB-FNO utilizes FNO module $\mathcal{G}_i(\cdot)$, parameterized by $\mathbf{W}_i$, to predict the above-mentioned residual $r_i$. 
The prediction result from the previous FNO is also adopted as input to $\mathcal{G}_i(\cdot)$ to ensure sufficient information is given for predicting the residual $r_i$. Hence, the input channel of the $i$-th FNO $\mathcal{G}_i$ is 2 times larger than that of the first FNO $\mathcal{G}_0$.
We can easily calculate the output of residual module $\mathcal{G}_i$ as
\begin{equation} \label{eq:residual_predict}
    \hat{r}_{i} = \mathcal{G}_i(x_{(i)} | \mathbf{W}_i), \quad x_{(i)} = [x, \hat{r}_{i-1}].
\end{equation}
Here $[, \ ]$ stands for concatenation operation. Therefore, the training objective for the $i$-th stage can be formulated as follows:
\begin{equation} \label{eq:loss_stage_i}
    \min_{(x, y) \in \mathcal{D}} \mathcal{L}_{\text{NRMSE}}(r_i, \hat{r}_i).
    % \min_{\mathbf{W}_i} \text{R-MSE}(r_i, \hat{r}_i)
\end{equation}

After finishing the training of the last module $\mathcal{G}_T$, all the preceding FNOs can inference as one ensemble, shown in Figure \ref{fig:boost}. The final prediction can be calculated as:
$\hat{y} = \sum^T_{i=0} \hat{r}_i$
Finally, the training process of SpecB-FNO is shown in Algorithm \ref{alg:specbfno} list in Appendix \ref{sec:appendix_algorithm}.

\section{Experiments}
\label{sec:exp}

% In this section, we conduct numerical experiments to validate the effectiveness of the SpecB-FNO method. We evaluate SpecB-FNO's performance across diverse PDE datasets, following previous work~\cite{FNO, NO-JMLR, FFNO, UNO, pdebench}. These datasets involve predicting PDE solutions from initial conditions and sequential PDE predictions. They contain PDEs with predominantly low-frequency information as well as those rich in high-frequency information. They also include tasks of single feature prediction and multiple coupled feature prediction.

In this section, we conduct numerical experiments to validate SpecB-FNO. 
We first describe the experimental setup in Section \ref{sec:exp_setup}.
Section \ref{sec:exp_effective} highlights SpecB-FNO's significant error reduction across various PDE datasets. 
In Section \ref{sec:exp_spectral}, we discuss SpecB-FNO's spectral performance, demonstrating its capability to address the Fourier parameterization bias and explaining the superior performance in Section \ref{sec:exp_effective}. 
Section \ref{sec:exp_efficiency} investigates the efficiency of SpecB-FNO and demonstrates that SpecB-FNO's effectiveness is not due to parameter increase.
% Section \ref{sec:exp_resolution} shows that SpecB-FNO enables FNO to learn high-resolution and high-frequency information from PDE data effectively.

\subsection{Experiment Description}
\label{sec:exp_setup}

\textbf{Datasets.} We conduct the evaluation on five datasets provided by previous research~\cite{FNO, pdebench}: (i) \& (ii) the incompressible Navier-Stokes equation for sequential prediction with $\nu$ = 1e-3 and $\nu$ = 1e-5, (iii) the steady-state Darcy flow equation for the initial condition to PDE solution prediction, (iv) the shallow water equation for sequential prediction, and (v) the diffusion-reaction equation for multi-feature sequential prediction. 
% The results of two coupled features (activator and inhibitor) in (v) are reported separately.
Details are introduced in Appendix \ref{sec:appendix_dataset}.

\textbf{Baselines.} To demonstrate the effectiveness of SpecB-FNO, we compare the following baselines with SpecB-FNO: (i) Conv-based surrogate models: ResNet~\cite{resnet}, U-Net~\cite{U-Net}, CNO~\cite{CNO} (ii) MLP-based surrogate model: DeepONet~\cite{deeponet}, (iii) Fourier-based surrogate models: FNO~\cite{FNO}, FFNO~\cite{FFNO}. Detailed descriptions are available in Section \ref{sec:appendix_baseline}.

% \textbf{Solo \textit{vs.} SpecB-FNO.} For all evaluations, \textbf{Solo} denotes training a solo model, while \textbf{SpecB-FNO} represents training an ensemble of two models sequentially.

\textbf{Metric and Significance.} Aligned with previous work~\cite{FNO, FFNO}, NRMSE in Eqn.~\eqref{eq:rmse} is adopted for evaluation. For all results, we report the mean $\pm$ std across three random seeds.

\textbf{Training and Evaluation Procedure. } For sequential PDE datasets, following previous work~\cite{FFNO}, teacher forcing is adopted during the training process. All models employ autoregressive prediction with one-step input and one-step output data. The NRMSE is averaged on the entire prediction sequence except for Darcy flow. % The evaluation begins by taking the 10th data in the sequence as input, predicting the remaining sequence.

\subsection{Effectiveness of SpecB-FNO}
\label{sec:exp_effective}

\begin{table}[!htbp]
\centering
\vspace{-5pt}
\caption{Error Comparison between SpecB-FNO and Baselines}   \label{tab:main}
% \resizebox{\textwidth}{!}{
\begin{tabular}{c|ccccc}
\hline
    \multirow{2}{*}{Model} & Darcy & \multicolumn{2}{c}{Navier-Stokes} & Shallow & Diffusion \\
\cline{3-4}
    % & & & $\nu$ = 1e-3 & $\nu$ = 1e-5 & & \\
    & flow & $\nu$ = 1e-3 & $\nu$ = 1e-5 & Water & Reaction \\
\hline
    DeepONet  & .0428$\pm$.0007 & .0716$\pm$.0018 & .2484$\pm$.0027 & .1576$\pm$.0216 & NaN \\
\hline
    ResNet    & .2455$\pm$.0011 & .9946$\pm$.2337 & .3926$\pm$.0007 & 1.501$\pm$.1519 & .0138$\pm$.0016 \\
    U-Net     & .0098$\pm$.0005 & .1105$\pm$.0547 & .1334$\pm$.0071 & 2.088$\pm$.2135 & .1160$\pm$.0068 \\
    CNO       & .0075$\pm$.0014 & .0512$\pm$.0017 & .1203$\pm$.0072 & .0326$\pm$.0021 & .0257$\pm$.0088 \\
\hline
    FNO       & \underline{.0067$\pm$.0001} & \underline{.0039$\pm$.0004} & \underline{.0576$\pm$.0004} & \underline{.0050$\pm$.0001} & .0190$\pm$.0003 \\
    FFNO      & .0096$\pm$.0001 & .0317$\pm$.0023 & .1499$\pm$.0219 & .0540$\pm$.0119 & \underline{.0072$\pm$.0001} \\
\hline
    SpecB-FNO & \textbf{.0036$\pm$.0002} & \textbf{.0014$\pm$.0001} & \textbf{.0351$\pm$.0018} & \textbf{.0004$\pm$.0002} & \textbf{.0066$\pm$.0003} \\
\hline
    Abs. Impr & .0031 & .0025 & .0225 & .0046 & .0006 \\
    Rel. Impr & 46.6\% & 63.3\% & 39.0\% & 92.5\% & 8.3\% \\
\hline
\end{tabular}
% }
\begin{tablenotes}
\footnotesize
\item[1] \textit{NaN} indicates that the experiment does not converge. The best-perform model and best-performed baseline are highlighted in \textbf{bold} and \underline{underline} respectively. \textit{Abs. Impr} and \textit{Rel. Impr} stands for absolute and relative improvement compared to best-peformed baselines, respectively.
\end{tablenotes}
\vspace{-5pt}
\end{table}

% \begin{table}[!htbp]
% \centering
% \vspace{-5pt}
% \caption{Error Comparison between SpecB-FNO and Baselines}   \label{tab:main}
% % \resizebox{\textwidth}{!}{
% \begin{tabular}{c|c|ccc}
% \hline
%     Category & Model & D.F. & N.S. ($\nu$ = 1e-3) & N.S. ($\nu$ = 1e-5) \\
% \hline
%     \multirow{1}{*}{MLP} 
%     & DeepONet & \\
% \hline
%     \multirow{3}{*}{Conv} 
%     & ResNet & x.xx $\pm$ 0.yy $\times 10^{-4}$ & x.xx $\pm$ 0.yy $\times 10^{-4}$ & x.xx $\pm$ 0.yy $\times 10^{-4}$ \\
%     & U-Net & \\
%     & CNO & \\
% \hline
%     \multirow{3}{*}{Fourier}
%     & FNO & \\
%     & FFNO & \\
%     & SpecB-FNO & \\
% \hline
%     Category & Model & S.W. & D.R. \\
% \hline
%     \multirow{1}{*}{MLP} 
%     & DeepONet & \\
% \hline
%     \multirow{3}{*}{Conv} 
%     & ResNet & x.xx $\pm$ 0.yy $\times 10^{-4}$ & x.xx $\pm$ 0.yy $\times 10^{-4}$  \\
%     & U-Net & \\
%     & CNO & \\
% \hline
%     \multirow{3}{*}{Fourier}
%     & FNO & \\
%     & FFNO & \\
%     & SpecB-FNO & \\
% \hline
% \end{tabular}
% % }
% \begin{tablenotes}
% \footnotesize
% \item[1] Here \textit{N.S.} denotes Navier-Stokes, \textit{D.F.} denotes Darcy-Flow, \textit{S.W.} denotes Shallow-Water, \textit{D.R.} denotes Diffusion-Reaction.
% \end{tablenotes}
% \vspace{-5pt}
% \end{table}

We compare the performance of SpecB-FNO with other baselines over the above-mentioned five datasets in Table \ref{tab:main} and make the following observations.
Firstly, SpecB-FNO constantly outperforms other surrogate models across all datasets, validating the effectiveness of spectral boosting. 
Secondly, the relative performance of surrogate models varies across different datasets. For example, while ResNet generally performs worse than FNO, it outperforms FNO on the diffusion-reaction equation. This dataset mainly contains local details and very few global features, making it naturally suited for ResNet with its local convolution kernels. Therefore, it's important to consider the physical and spectral properties of a specific PDE when choosing surrogate models.
%Secondly, the best-performed baselines differ on different datasets. CNO performs the best on the Darcy flow dataset, FNO performs the best on the Naiver-Strokes and shallow water datasets, while FFNO outperforms the rest on the Diffusion Reaction dataset. Such a phenomenon is likely caused by the diverse physical and spectral properties of different PDEs.
% Such a phenomenon is likely caused by the diverse physical and statistical nature of different partial derivate equations.
Thirdly, it can be observed that specifically designed neural operator learning surrogate models, such as CNO, FNO, or FFNO, generally outperform other surrogate models adapted from computer vision tasks, such as U-Net and ResNet. This empirically reflects the distinction between PDE tasks and classic CV tasks, highlighting the necessity of customized-designed surrogate models.
Lastly, DeepONet, as an MLP-based surrogate model, is generally outperformed by the latest Conv-based and Fourier-based surrogate models, such as FNO and CNO. % This reflects the importance of utilizing various convolution kernels, spatial or spectral, in surrogate models, as these kernels can effectively capture global and local features.
This highlights the importance of using convolution kernels parameterized in either the spatial or Fourier domain to capture both global and local features when learning PDEs on grid data.

It is worth mentioning that in Table \ref{tab:main} SpecB-FNO achieves optimal performance with larger kernels than FNO in all cases, detailed in Appendix ~\ref{sec:appendix_hyperparam}. FNO typically performs best with a relatively small truncation frequency, consistent with previous research~\cite{FNO, FFNO, G-FNO, DAFNO, iFNO}. In contrast, SpecB-FNO performs best with a significantly larger frequency mode. Particularly for the Navier-Stokes ($\nu$ = 1e-5), shallow water, and diffusion-reaction datasets, SpecB-FNO achieves optimal performance with a Fourier kernel that preserves all frequency modes within the target data resolution. This indicates that SpecB-FNO addresses the bottleneck of FNO's ineffectiveness with large Fourier kernels.

\subsection{Spectral Analysis of SpecB-FNO}
\label{sec:exp_spectral}
% \vspace{-15pt}

This section presents a spectral analysis of SpecB-FNO, showing that it effectively mitigates the Fourier parameterization bias and enables FNO to better utilize parameters across all frequencies within its Fourier kernels rather than focusing only on the dominant frequencies.

Figure \ref{fig:spectral} illustrates the NMSE spectrum of FNO and SpecB-FNO. SpecB-FNO provides the greatest relative improvements below the truncation frequency, particularly at the non-dominant frequencies of the target data. For example, for the PDEs in Figures \ref{fig:spectral-darcy} to \ref{fig:spectral-shallow}, where dominant frequencies are near modes 0 and 1, SpecB-FNO mainly enhances FNO's performance at higher frequencies within the truncation frequency. For the PDE in Figure \ref{fig:spectral-diffusion-a}, with dominant frequencies around mode 10, the most significant improvement from SpecB-FNO occurs on either side of the Fourier kernel. For the PDE in Figure \ref{fig:spectral-diffusion-i}, with dominant frequencies around mode 5, the greatest improvements from SpecB-FNO are seen at higher frequencies within the Fourier kernel. These observations indicate that SpecB-FNO effectively improves FNO's performance on non-dominant frequencies.

\paragraph{SpecB-FNO performance on larger Fourier kernels}
\label{sec:large_kernel}

\begin{figure}
\centering
\vspace{1pt}
\begin{subfigure}{0.32\textwidth}
    \includegraphics[width=\textwidth]{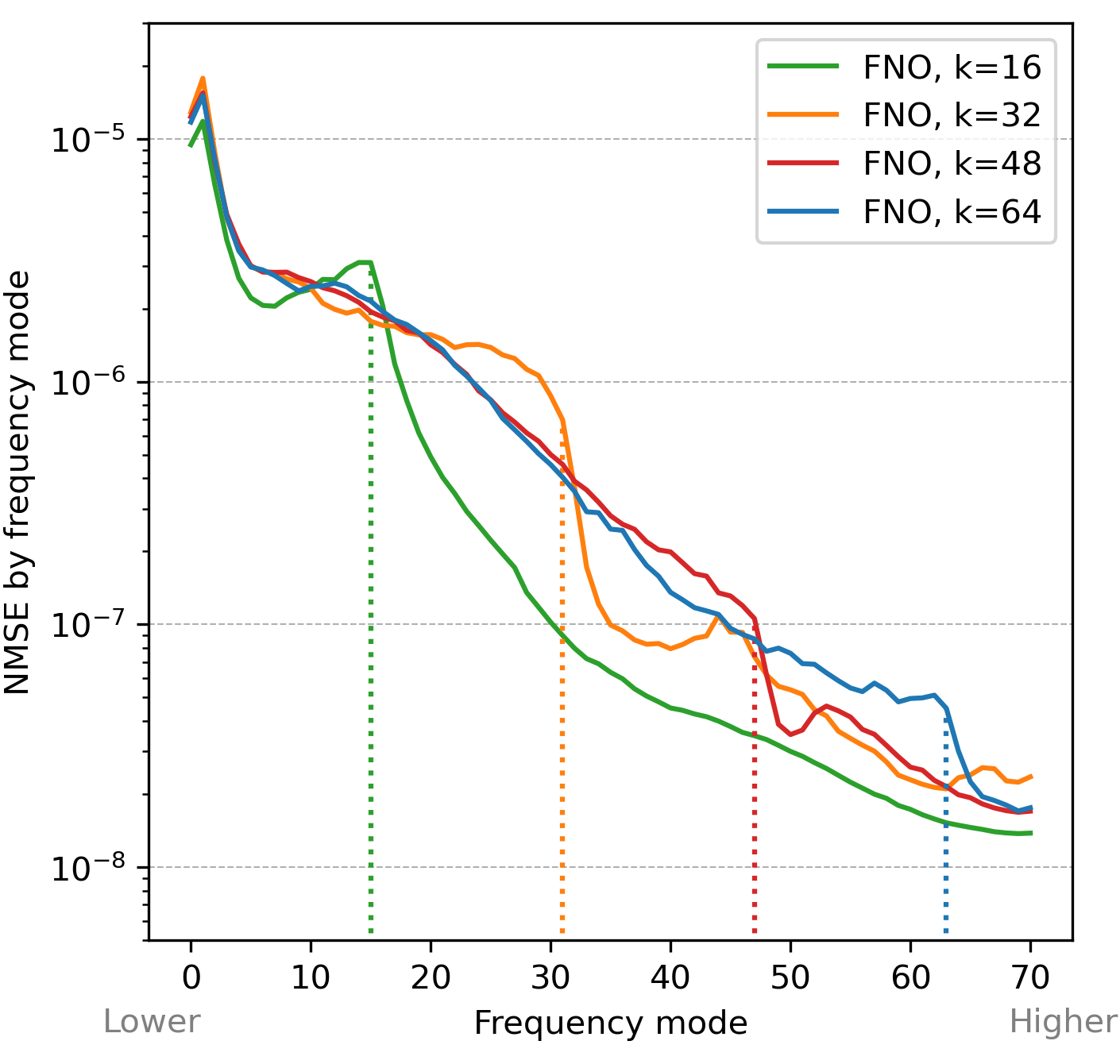}
    \caption{Initial spectral performance.}
    \label{fig:spectral-darcy1}
\end{subfigure}
\hfill
\begin{subfigure}{0.32\textwidth}
    \includegraphics[width=\textwidth]{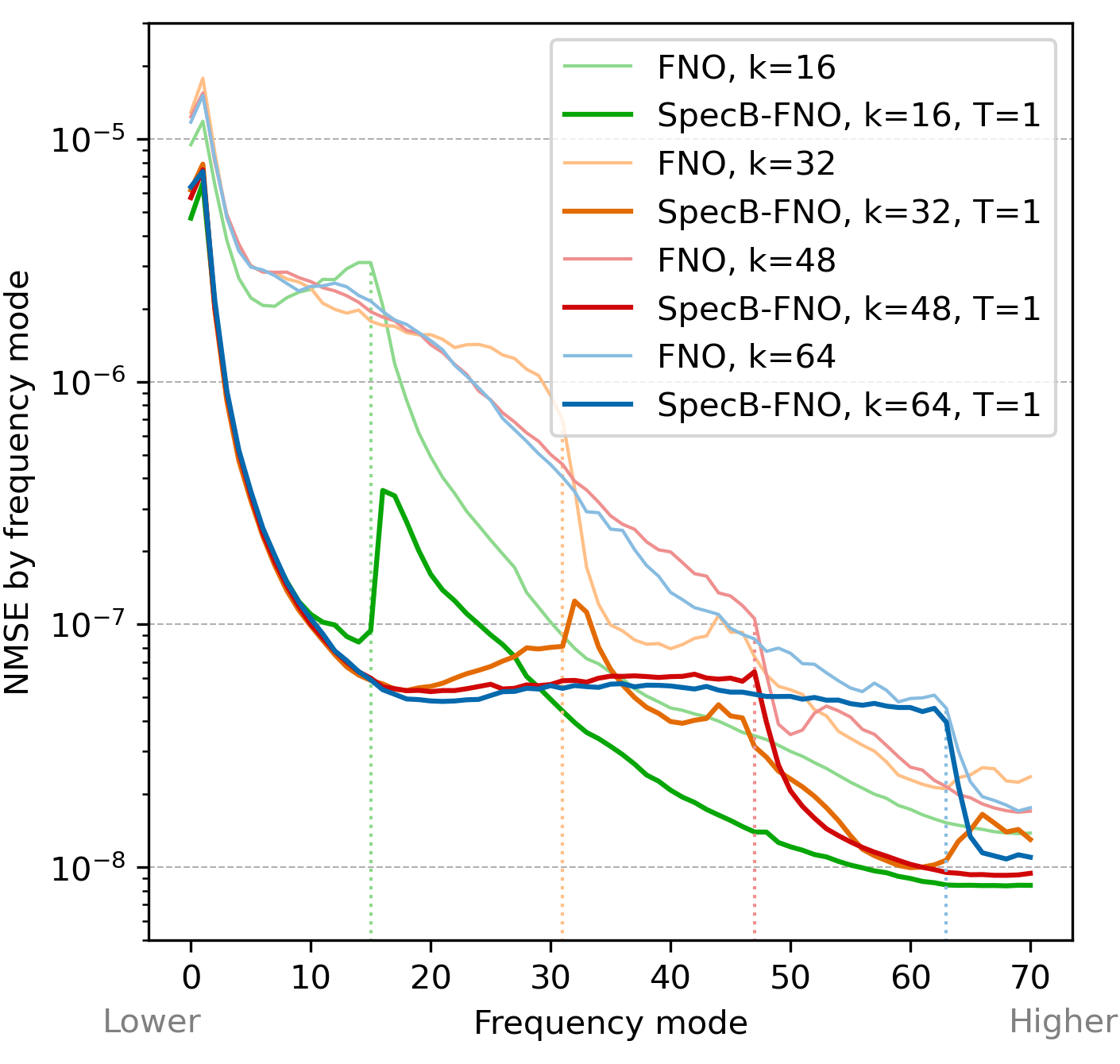}
    \caption{Improvements with T=1.}
    \label{fig:spectral-darcy2}
\end{subfigure}
\hfill
\begin{subfigure}{0.32\textwidth}
    \includegraphics[width=\textwidth]{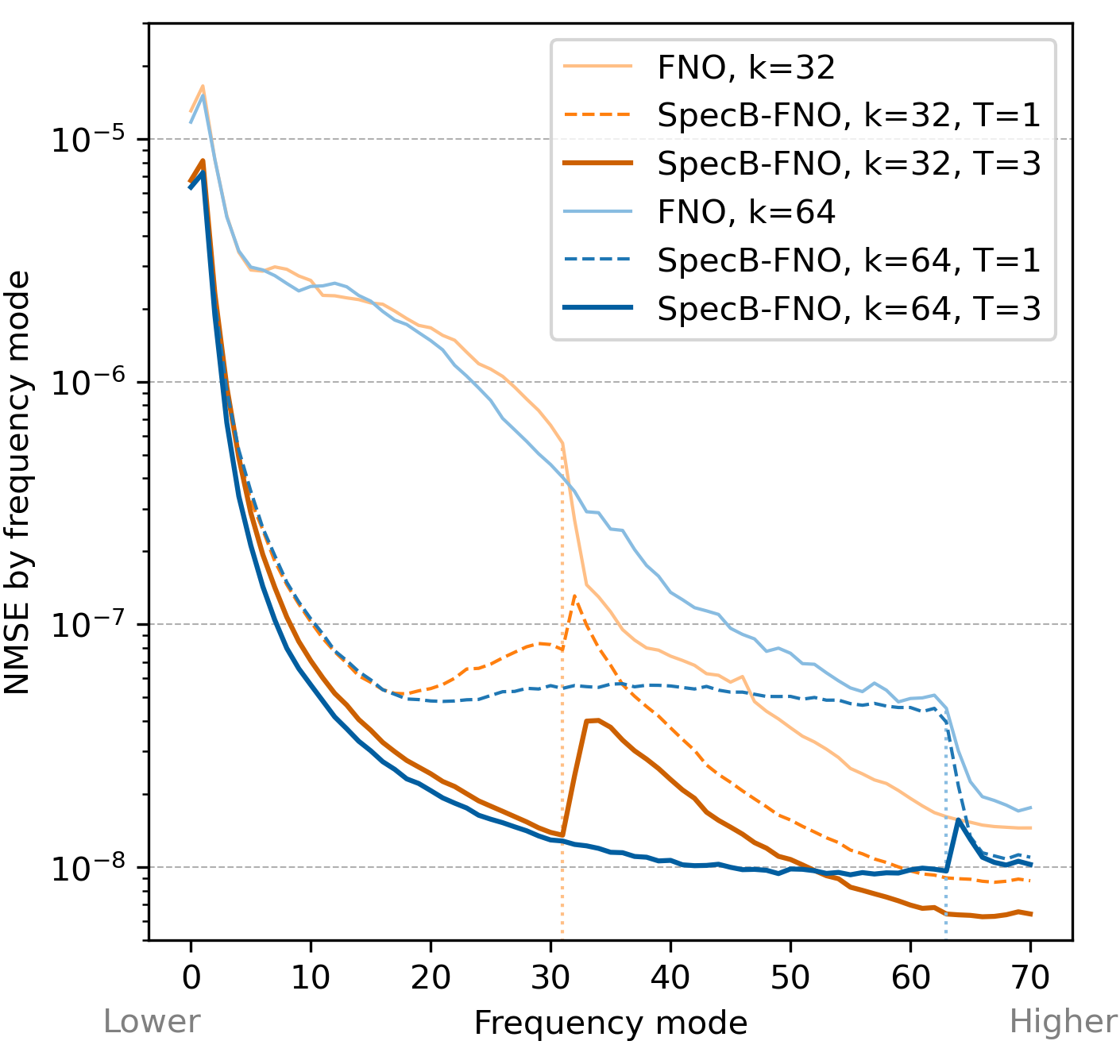}
    \caption{Improvements with T=3.}
    \label{fig:spectral-darcy3}
\end{subfigure}
\vspace{-5pt}
\caption{NMSE spectrums on Darcy flow with different stages of SpecB-FNO. The truncation frequency, k, is marked with a dotted line. In the initial stage, SpecB-FNO collapses to FNO.}
\vspace{-10pt}
\label{fig:spectral-darcy-all}
\end{figure}

In FNO, larger Fourier kernels can exhibit a stronger Fourier parameterization bias, which is harder to address and may require more stages of spectral boosting. This occurs because, once FNO's Fourier kernel already covers the dominant frequencies, further increasing the truncation frequency only includes more non-dominant frequencies, amplifying the Fourier parameterization bias. We demonstrate SpecB-FNO's performance with larger Fourier kernels in Figure \ref{fig:spectral-darcy-all} on the Darcy flow dataset.
% When FNO's Fourier kernel already covers the dominant frequencies, further increasing the truncation frequency will only include more non-dominant frequencies to the Fourier kernel. Consequently, larger Fourier kernels can exhibit a stronger Fourier parameterization bias, which is harder to address and may require more stages of SpecB-FNO. We will demonstrate this phenomenon in Figure \ref{fig:spectral-darcy-all} over the Darcy flow dataset.

In Figure \ref{fig:spectral-darcy2}, we show the spectral performance of FNO with different truncation frequencies after one stage of spectral boosting. With larger Fourier kernels, particularly FNO with $k=64$, improvements in the least dominant frequencies around mode 64 are very limited. Given that (i) frequencies around mode 64 can't be well learned by a solo FNO with $k=64$ due to Fourier parameterization bias, and (ii) performance around mode 64 does not significantly improve after one stage of spectral boosting, we can infer that there is still room to improve the performance near mode 64. In Figure \ref{fig:spectral-darcy3}, we show the results after two more stages of spectral boosting. The performance near mode 64 is indeed further improved. This indicates that FNO's Fourier parameterization bias with larger Fourier kernels can be harder to address and may require more stages of spectral boosting. 
% {\color{red}For this dataset, additional stages of SpecB-FNO do not significantly enhance FNO's performance further.}

Another interesting observation from Figure \ref{fig:spectral-darcy3} is that, after being sufficiently optimized by spectral boosting, the spectral performances of FNO-32 and FNO-64 at lower frequencies converge to a similar level. This occurs because increasing FNO's truncation frequency from 32 to 64 only adds Fourier parameters for learning frequencies above mode 32. The parameters for learning frequencies below mode 32 remain unchanged. Therefore, if FNO can fully utilize its Fourier kernels, increasing the truncation frequency will primarily improve high-frequency performance rather than low-frequency.

\subsection{Ablation Study on Efficiency}
\label{sec:exp_efficiency}

\paragraph{Ablation on Parameter Size}

% As discussed in Section \ref{sec:method_specboost}, SpecB-FNO utilizes multiple FNOs to iteratively learn the residual of the previous one, yielding significant performance increases. In this section, we compare SpecB-FNO with FNOs given roughly the same amount of model parameters. We first enlarge FNO on two different dimensions: layer and channel. Given that the number of parameters increases exponentially with the channel, we enlarge FNO by layer and channel by 2 $\times$ and 1.5 $\times$, respectively, and compare their prediction accuracy with SpecB-FNO (T=2).
As discussed in Section \ref{sec:specbfno-specb}, SpecB-FNO utilizes FNO modules to iteratively learn the residuals of the previous ones, resulting in significant performance improvements. In this section, we compare SpecB-FNO with FNOs, which have roughly the same amount of parameters, to show that SpecB-FNO's superiority is not due to parameter increase.
Since SpecB-FNO with $T=2$ contains twice the parameters of one FNO module, we increase FNO's parameters by increasing its hidden channels by 1.5$\times$ or its layers by 2$\times$. All other hyperparameters of models in Table \ref{tab:efficiency} are the same, including the truncation frequency.

\begin{table}[!htbp]
\centering
\vspace{-5pt}
\caption{Efficiency Comparision between SpecB-FNO and baselines.}  \label{tab:efficiency}
% \resizebox{\textwidth}{!}{
\begin{tabular}{c|ccccc}
\hline
    \multirow{2}{*}{Model} & Darcy & \multicolumn{2}{c}{Navier-Stokes} & shallow & diffusion- \\
\cline{3-4}
    % & & & $\nu$ = 1e-3 & $\nu$ = 1e-5 & & \\
    & flow & $\nu$ = 1e-3 & $\nu$ = 1e-5 & water & reaction \\
\hline
    FNO       & .0092$\pm$.0001 & \underline{.0047$\pm$.0002} & .0603$\pm$.0007 & .0050$\pm$.0001 & .0190$\pm$.0004 \\
    FNO-c     & \underline{.0077$\pm$.0003} & \underline{.0047$\pm$.0001} & \underline{.0594$\pm$.0007} & .0044$\pm$.0004 & .0229$\pm$.0025   \\
    FNO-l     & .0082$\pm$.0002 & .0230$\pm$.0001 & .0602$\pm$.0004 & \underline{.0043$\pm$.0002} & \underline{.0169$\pm$.0006} \\
\hline
    Param. Impr & 16.3\% & 0.0\% & 1.5\% & 14.0\% & 11.1\% \\
\hline
\hline
    SpecB-FNO & \textbf{.0039$\pm$.0003} & \textbf{.0014$\pm$.0001} & \textbf{.0351$\pm$.0018} & \textbf{.0014$\pm$.0001} & \textbf{.0066$\pm$.0003} \\
\hline
    SpecB. Impr & 57.6\% & 70.2\% & 41.8\% & 72.0\% & 65.3\% \\
\hline
\end{tabular}
% }
\begin{tablenotes}
\footnotesize
\item[1] \textit{FNO-c} and \textit{FNO-l} refers to enlarging FNO models by increasing channel and layer by 1.5 $\times$ and 2 $\times$. The best-perform model and best-performed baseline are highlighted in \textbf{bold} and \underline{underline} respectively. \textit{Param. Impr} and \textit{SpecB. Impr} stands for relative improvement bought by parameter increase and spectral boosting. \textit{Param. Impr} equals the maximum improvement of FNO-c or FNO-l than FNO, while \textit{SpecB. Impr} represents the improvement of SpecB-FNO than FNO.
\end{tablenotes}
\vspace{-5pt}
\end{table}

We report the ablation result in Table \ref{tab:efficiency}. We can easily observe that with the same amount of parameters, the performance increase bought by spectral boosting is much larger than that bought by parameter increase. Such an observation indicates that the error reduction of SpecB-FNO is mainly caused by specific designs tackling Fourier parameterization bias instead of parameter increase.

\paragraph{Ablation on Training Efficiency and Memory Utility} 
Training efficiency and GPU utility are important features affecting SpecB-FNO's usage in the real world, especially for large PDE datasets with high resolutions. We report these features in Table \ref{tab:nslayer} on the Navier-Stokes ($\nu$ = 1e-5) dataset. We can easily observe that SpecB-FNO requires less GPU memory than FNO-c and FNO-l, as it only trains part of the parameters for each stage. Such a memory-efficient property enables the training of large models. On the other hand, although trained iteratively, SpecB-FNO exhibits roughly the same amount of training time compared to FNO-l and FNO-c.

\begin{table}[!htbp]
\centering
\vspace{-5pt}
\caption{Efficiency Comparision between SpecB-FNO and baselines on Navier-Stokes ($\nu$ = 1e-5).}  \label{tab:nslayer}
% \resizebox{\textwidth}{!}{
\begin{tabular}{c|ccccc}
\hline
     Dataset & FNO & FNO-l & FNO-c & SpecB-FNO (T=1) \\
     \hline
     Param Count (million) & 328 & 656 & 738 & 656 & \\
     Train Max Memory (MB/instance)  & 341 & 613 & 536 & 400 \\
     Total Training Time (hour) & 1.81 & 3.54 & 3.38 & 3.63 \\
\hline
\end{tabular}
% }
\footnotesize
\begin{tablenotes}
\item[1] \textit{FNO-c} and \textit{FNO-l} refers to enlarging FNO models by increasing channels and layers by 1.5 $\times$ and 2 $\times$.
\end{tablenotes}
\vspace{-5pt}
\end{table}

\section{Related Work}

\subsection{Neural Networks for Solving PDEs}
Recognized for their exceptional approximation capabilities, neural networks have emerged as a promising tool for tackling PDEs. 
Physics-Informed Neural Networks (PINNs)~\cite{pinn} leverage neural networks to fit the PDE solutions in a temporal and spatial range while adhering to PDE constraints. 
On the other hand, the operator learning paradigm, such as DeepONet \cite{deeponet}, neural operators~\cite{NO-JMLR}, spectral neural operator~\cite{SNO}, LOCA~\cite{LOCA}, message passing neural PDE solvers~\cite{MP-PDE} and transformer-based models~\cite{Galerkin}, offers alternative approaches by employing neural networks to fit the complex operators in solving PDEs, directly mapping input functions to their target functions. 
Classic convolution-based models such as ResNet~\cite{resnet} or U-Net~\cite{U-Net} have also been adapted to solve PDEs as surrogate models. Researchers also propose adaptations~\cite{CNO, gupta2022towards} upon these classic models.

Among the neural operators, FNO \cite{FNO} incorporates the Fast Fourier Transform (FFT) \cite{cochran1967fast} in its network architecture, achieving both advantageous efficiency and prediction accuracy. Its universal proximity is also proven~\cite{PsiFNO}.
As a resolution-invariant model, FNO trained on low-resolution data can be directly applied to infer on high-resolution data.
Notable efforts have been made to enhance the performance of FNO from various aspects \cite{FFNO, T1, UNO, gupta2022towards, BOON, Clifford, G-FNO, wang2024beno, he2023mgno, tu2023guaranteed}.
% Notable efforts have been made to enhance the performance of FNO from various aspects, such as factorizing Fourier kernels \cite{FFNO}, minimizing Fourier transform times \cite{T1}, reducing memory consumption \cite{UNO}, applying physics constraints \cite{BOON}, optimizing vector data utilization \cite{Clifford}, and achieving rotation and translation invariance \cite{G-FNO}. 
Several studies aim to improve FNO's effectiveness in solving PDEs with distinctive properties, including coupled PDEs \cite{CMWNO}, physics-constrained~\cite{PINO}, inverse problems for PDEs \cite{NIO}, and steady-state PDEs \cite{FNO-DEQ}. 
Since FNO relies on Fourier transform on regular meshed grids, broad work focuses on enabling FNO to process various data formats, including irregular grids \cite{DAFNO}, spherical coordinates \cite{SFNO}, cloud points \cite{GINO}, and general geometries \cite{li2023fourier, CORAL}.

Despite recent advances, FNO's ineffectiveness with large Fourier kernels has not been sufficiently discussed. Previous research adopts small Fourier kernels~\cite{iFNO,FNO,FFNO}, thereby restricting FNO's ability to learn from complex PDE data and further enhance its accuracy. SpecB-FNO aims to investigate and mitigate such limitations.

%In this work, the proposed SpecB-FNO framework doesn't involve altering the internal architecture of neural operators. 
% Consequently, SpecB-FNO is orthogonal to most existing methods, as it imposes no requirements on the model architecture or data formats.

% Despite the recent advances, the low-frequency bias and its impact have not been explored in neural operators.
% In this work, in contrast to the majority of neural operator variants, the proposed approach doesn't involve altering the internal architecture of neural operators. 
% Instead, by examining FNO's frequency properties, we suggest that a direct ensemble of two FNOs has the potential to predict high frequencies more accurately. 

\subsection{Spectral Properties for Neural Networks}

\textbf{Low-frequency bias.} It has been observed that during the training process, neural networks employing the ReLU activation function tend to first learn low frequencies in data and progress more slowly in learning high frequencies \cite{bias1, bias2}. This characteristic diverges from traditional numerical solvers, which typically converge on high frequencies first. 

In this study, we identify a unique spectral property: Fourier parameterization bias. Unlike the typical low-frequency bias in general neural networks, Fourier parameterization bias refers to a preference for the dominant frequencies in the target PDE data, which are not necessarily low frequencies.

% A recent study claims that FNO exhibits the typical low-frequency bias found in general neural networks. However, we believe this is inaccurate because it overlooks FNO's distinct spectral properties below and beyond the truncation frequency. In fact, a close examination of Figure 7 in that paper reveals evidence supporting our observations: (i) FNO exhibits different spectral performance below and beyond its truncation frequency, and (ii) FNO is biased toward the dominant frequencies in the target data, not the lowest frequencies. However, the authors did not explicitly highlight these points and instead treated FNO's spectral performance as the typical low-frequency bias.

\textbf{Spectral performance of FNO.} In the existing literature, the spectral performance of FNO has not been widely explored. One study~\cite{iFNO} observes high-frequency noise in large Fourier kernels but does not explain its reason. Instead of making large Fourier kernels more effective, it focuses on automatically selecting small Fourier kernels based on the target PDE data. Another study~\cite{HANO} claims that FNO exhibits low-frequency bias and proposes a hierarchical attention neural operator (HANO) to address this issue. Our work differs from theirs because (i) HANO does not address FNO's limitations with large Fourier kernels, and (ii) HANO overlooks FNO's unique spectral performance and treats it as the typical low-frequency bias.

\section{Conclusion}
\label{sec:conclusion}

% This study provides empirical evidence from a spectral perspective to elucidate the superior performance of FNO over CNNs. The pivotal factor lies in FNO's exceptional capability to capture low-frequency information. Building on this insight, we introduce SpecBoost as a solution to mitigate the low-frequency bias in FNO caused by global Fourier filters. The results across different PDE tasks underscore the efficacy and efficiency of the SpecBoost framework.

% In this paper, we aim to tackle one limitation of FNO: FNO is ineffective with large Fourier kernels that parameterize more frequencies. By spectral analysis, we empirically conclude the Fourier parameterization bias that convolution kernels parameterized in spectral space exhibit a stronger bias toward the dominating frequencies than spatial space. To mitigate such a bias, we propose SpecB-FNO, which adopts additional FNO modules iteratively by learning from the previous ones' residuals. Empirical experiments on various PDE applications witness error reduction up to 93\%.

In this paper, we elucidate and address FNO's ineffectiveness with large Fourier kernels. Through spectral analysis, we identify a unique Fourier parameterization bias in FNO: convolution kernels parameterized in the Fourier domain exhibit a stronger bias toward the dominant frequencies in the target data compared to those parameterized in the spatial domain. % We propose SpecB-FNO to mitigate this bias and demonstrate that if parameters in Fourier kernels are fully utilized, larger kernels can significantly improve FNO's accuracy. 
% We propose SpecB-FNO to mitigate this bias and demonstrate that fully utilizing parameters in large Fourier kernels can significantly improve FNO's accuracy, with an observed error reduction of up to 93\%.
We propose SpecB-FNO to mitigate this bias and show that when parameters in Fourier kernels are fully utilized, larger kernels can significantly improve FNO's accuracy, with an average 50\% reduction in error.

\newpage
\bibliographystyle{plain}
\bibliography{neurips_2024.bib}

%%%%%%%%%%%%%%%%%%%%%%%%%%%%%%%%%%%%%%%%%%%%%%%%%%%%%%%%%%%%

\newpage
\appendix

\section{Formulating Different Versions of Fourier Layer}
\label{sec:appendix_fno}

The key architecture of FNO is centered around its Fourier layer. In the main paper, we adopt the latest implementation from the author's official repository \footnote{https://github.com/neuraloperator/neuraloperator/}. 

\begin{figure}[!htbp]
    \centering
    \includegraphics[width=0.7\columnwidth]{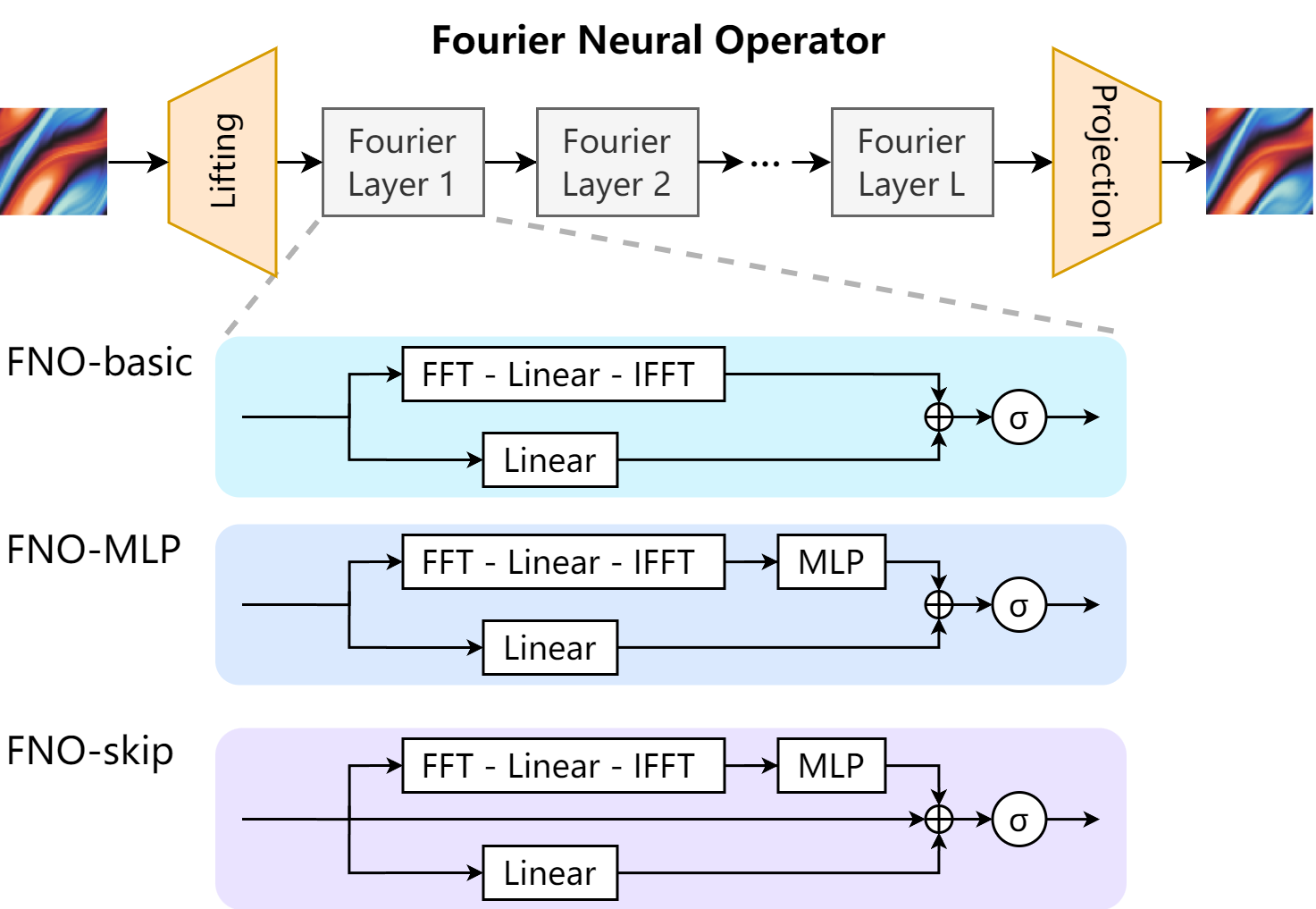}
    \vspace{-5pt}
    \caption{FNO architecture and designs for Fourier layers.}
    \vspace{1pt}
    \label{fig:fno_version}
\end{figure}

In the original paper, the basic Fourier layer consists of a pixel-wise linear transformation $\phi$, and an integral kernel operator $\mathcal{K}$, denoted as:
\begin{equation}
\mathcal{H}^{basic}(x) =  \sigma \left(\phi(x) + \mathcal{K}(x) \right),
\end{equation}
with $\sigma$ as the nonlinear activation function.
The integral kernel operator $\mathcal{K}$ undergoes a sequential process involving three operations: Fast Fourier Transformation (FFT)~\cite{cochran1967fast}, spectral linear transformation, and inverse FFT. The primary parameters of FNO are located in the spectral linear transformation. 
% Hence, FNO truncates high-frequency modes in each Fourier layer to avoid introducing extensive parameters. 
Hence, FNO truncates high-frequency modes in each Fourier layer to decrease the parameter size and also prevent high-frequency noise. 
These truncated frequency modes can encompass rich spectrum information, especially for high-resolution inputs.

The authors of~\cite{FNO} have also introduced alternative configurations for Fourier layers in their publicly available code. 
One adjustment involves incorporating a pixel-wise MLP, denoted as $\mathcal{M}$, after the kernel operator $\mathcal{K}$:
\begin{equation}
\label{eq:fno-m}
\mathcal{H}^{MLP}(x) =  \sigma \left(\phi(x) + \mathcal{M}(\mathcal{K}(x)) \right).
\end{equation}
The last modification to FNO involves including skip connections, which are commonly employed in training deep CNNs \cite{resnet}. Similar to our main paper, this version of the Fourier layer can be formulated as follows:
\begin{equation}
\label{eq:fno-s}
\mathcal{H}^{skip}(x) =  \sigma \left(x + \phi(x) + \mathcal{M}(\mathcal{K}(x)) \right).
\end{equation}
It's shown that employing skip connections to Fourier layers enables the training of a deeper FNO~\cite{FFNO}. We choose the FNO-skip setting for FNO for all experiments in the main paper and abbreviate $\mathcal{H}^{skip}()$ as $\mathcal{H}()$. All these FNO versions can be visualized in Figure ~\ref{fig:fno_version}.

\section{NMSE Spectrum Computation}
\label{sec:nmse_spectrum_computation}

Here, we describe how to compute the NMSE spectrum used in our paper. First, we obtain the normalized prediction residual (the normalized difference between the target and prediction) for all model predictions on the test set. Here, normalizing means dividing all pixels in the 2D data by a scalar that ensures the mean energy in the target data is 1.

For each normalized prediction residual, we use FFT to convert it to the Fourier domain and shift the lowest frequency to the center of the 2D spectrum. We then compute the pixel-wise energy of this 2D spectrum and divide it by the total resolution of the spectrum \textit{twice}. After the first division, the sum of energy in the spectrum equals the \textit{sum of energy} in the normalized prediction residual. After the second division, the sum of energy in the spectrum equals the \textit{average energy} in the normalized prediction residual, which is the NMSE.

Next, we redistribute the energy of the 2D spectrum into 1D with respect to frequency modes. Mode 0 contains the energy of the center pixel in the spectrum. Mode 1 contains the energy of the 8 pixels surrounding the center pixel. Mode 2 contains the energy of the 16 pixels surrounding the previous 8 pixels, and so on. This process yields the NMSE spectrum for one testing sample. The final NMSE spectrum is the average NMSE spectrum across all test data.

\section{Pseudo Algorithm}
\label{sec:appendix_algorithm}

Here, we list the pseudo algorithm of SpecB-FNO in Algorithm \ref{alg:specbfno}.
\begin{algorithm}
	\caption{Training Process of SpecB-FNO} 
    \label{alg:specbfno}
	\begin{algorithmic}[1]
		\Require training set $\mathcal{D}$, residual learning iterations $T$
        \Ensure model parameters set $\mathbf{W} = \{\mathbf{W}_i\}$
        \State Initialize the parameter set $\mathbf{W}$ as empty set $\phi$
        \State Train the initial FNO model $\mathcal{G}_0$ given Eq. \ref{eq:pde_obj} and obtain parameter $\mathbf{W}_0$.
        \State Add parameter $\mathbf{W}_0$ to the final parameter set $\mathbf{W}$
        \For{$i$ = $1$, $\cdots$, $T$}
            \While{not converge}
                \State Sample mini-batch $\mathcal{B} = \{x, y\}$ from training set $\mathcal{D}$
                \State Calulate label $r_i$ given Eq. \ref{eq:residual_label}
                \State Calulate prediction $\hat{r}_i$ given Eq. \ref{eq:residual_predict}
                \State Update model parameter $\mathbf{W}_i$ using gradients $\nabla_{\mathbf{W}_i} \mathcal{L}_{\text{MSE}}(r_i, \hat{r}_i|\mathbf{W}_i, \mathbf{W})$
            \EndWhile
            \State Add parameter $\mathbf{W}_i$ to the final parameter set $\mathbf{W}$
        \EndFor
	\end{algorithmic}
\end{algorithm}

\section{Detailed Experimental Setup}

\subsection{Datasets}
\label{sec:appendix_dataset}

\paragraph{Navier-Stokes equation~\cite{FNO}.} As a fundamental PDE in fluid dynamics, the Navier-Stokes equation finds significance in diverse applications, including weather forecasting and aerospace engineering. Here, we consider the 2D incompressible Navier-Stokes dataset for viscosity following \cite{FNO}:
\begin{equation}\label{eq:ns}
\begin{aligned}
    \partial_t w(x,t) + u(x,t) \cdot \nabla w(x,t) &= \nu \Delta w(x,t) + f(x) ,\\
    \nabla \cdot u(x,t) &= 0 ,\\
    w(x,0) &= w_0(x) .
\end{aligned}
\end{equation}
The equation involves the viscosity field $w(x,t) \in \mathbb{R}$, with an initial value of $w_0(x)$, while $u\in \mathbb{R}^2$ represents the velocity field. The solution domain spans $x \in (0,1)^2$, $t \in \{1, 2, \dots, T\}$. The forcing function is represented by $f(x)$. The viscosity coefficient, $\nu$, quantifies a fluid's resistance to deformation or flow. The dataset comprises experiments with two viscosity coefficients: $\nu$ = 1e-3 and 1e-5, corresponding to sequence lengths $T$ of 50 and 20, respectively. For smaller $\nu$ values, the flow field exhibits increased chaos and contains more high-frequency information.

The prediction task involves using the initial ten viscosity fields in the sequence to predict the remaining ones. The viscosity field resolution is 64 $\times$ 64. For all viscosities, we use 1000 sequences for training and 200 for testing. No data augmentation approach is applied.

\paragraph{Darcy flow equation~\cite{FNO}.} Consider the 2D steady-state Darcy flow equation following \cite{FNO}:
\begin{equation}\label{eq:darcy}
\begin{aligned}
    - \nabla \cdot(a(x) \nabla u(x)) &= f(x), &x\in (0, 1)^2 \\
    u(x) &= 0, &x\in \partial(0, 1)^2
\end{aligned}
\end{equation}
where $a(x)$ is the diffusion coefficient and $f(x)$ is the forcing function. The goal is to use coefficient $a(x)$ to predict the solution $u(x)$ directly. The dataset includes diffusion coefficients and corresponding solutions at a resolution of 421 $\times$ 421. Datasets at smaller resolutions are derived through downsampling.

A total of 2048 samples are provided. We use 1800 samples for training and 248 for testing. The training and testing resolution is 141 $\times$ 141. Training data is augmented through flipping and rotations at 90, 180, and 270 degrees. 

\paragraph{Shallow water equation~\cite{pdebench}.} The shallow water equations, derived from the general Navier-Stokes equations, present a suitable framework for modeling free-surface flow problems. In 2D, these come in the form of the following system of hyperbolic PDEs,

\begin{equation}
\begin{aligned}
\partial_t h+\partial_x h u+\partial_y h v & =0 \\
\partial_t h u+\partial_x\left(u^2 h+\frac{1}{2} g_r h^2\right) & =-g_r h \partial_x b \\
\partial_t h v+\partial_y\left(v^2 h+\frac{1}{2} g_r h^2\right) & =-g_r h \partial_y b
\end{aligned}
\end{equation}

with $u$, $v$ being the velocities in the horizontal and vertical direction, $h$ describing the water depth, and $b$ describing a spatially varying bathymetry. $hu$, $hv$ can be interpreted as the directional momentum components and $g_r$ describes the gravitational acceleration.

A total of 1000 sequences are provided, each containing 101 continuous time steps with a PDE data resolution of 128 $\times$ 128. To reduce the data size for faster training, we retain 1 time step out of every 5, resulting in sequences with 21 time steps. For each sequence, the task is to use the first time step as input and predict the remaining 20 time steps autoregressively. We use 800 sequences for training and 200 sequences for testing. No data augmentation approach is applied.

\paragraph{Diffusion-reaction~\cite{pdebench}.} The diffusion-reaction dataset contains non-linearly coupled variables, namely the activator $u = u(t, x, y)$ and the inhibitor $v = v(t, x, y)$. The equation is written as

\begin{equation}
\partial_t u=D_u \partial_{x x} u+D_u \partial_{y y} u+R_u, \quad \partial_t v=D_v \partial_{x x} v+D_v \partial_{y y} v+R_v,
\end{equation}

where $Du$ and $Dv$ are the diffusion coefficient for the activator and inhibitor, respectively, $Ru = Ru(u, v)$ and $Rv = Rv (u, v)$ are the activator and inhibitor reaction function determined by the Fitzhugh-Nagumo equation. The domain of the simulation includes $x \in (-1, 1)$, $y \in (-1, 1)$, $t \in (0, 5]$. This equation is applicable most prominently for modeling biological pattern formation.

A total of 1000 sequences are provided, each containing 101 continuous time steps with two features at a resolution of 128 $\times$ 128. To reduce data size for faster training, we retain only a sequence of length 11, starting from time step 10. We do not start from zero because the initial state of the diffusion-reaction equation resembles high-frequency noise, which cannot be captured by the baseline models. For each sequence, the task is to use the two features at the first time step as inputs and predict the two features at the remaining 10 time steps autoregressively. We use 800 sequences for training and 200 for testing. Training data is augmented through flipping and rotations at 90, 180, and 270 degrees. 

\subsection{Baseline Description}
\label{sec:appendix_baseline}

In this section, we introduce the baselines adopted in Section \ref{sec:exp} as follows:
\begin{itemize}[topsep=0pt,noitemsep,nolistsep,leftmargin=*]
    \item ResNet~\cite{resnet} is a convolution neural network. It addresses the problem of vanishing and exploding gradients with residual connections. ResNet is a widely adopted baseline in PDE prediction~\cite{FNO, CNO, gupta2022towards, HANO}.
    \item U-Net~\cite{U-Net} is a convolutional neural network (CNN) initially designed for image segmentation tasks. It first gradually reduces the image size by the encoder, then increases the size by the decoder. Skip-connection is adopted between layers. U-Net is a widely adopted baseline in PDE prediction~\cite{FNO, DAFNO, CNO, gupta2022towards, HANO}
    \item DeepONet~\cite{deeponet}, named deep operator network, is proposed to learn operators from a small dataset. It consists of two sub-networks, one for encoding the input function at a fixed number of sensors and another for encoding the locations for the output functions. Our implementation of DeepONet is adapted from the official implementation\footnote{https://github.com/lululxvi/deeponet}.
    \item FNO~\cite{FNO} is a deep learning approach that combines neural networks with the Fourier transform to solve PDEs. Notably, we use a more recent version of FNO from the PyTorch neural operator library\footnote{https://github.com/neuraloperator/neuraloperator/}, incorporating MLP and skip connections, detailed in Appendix ~\ref{sec:appendix_fno}. This version is more advanced than the original FNO described in its initial paper~\cite{FNO}.
    \item FFNO~\cite{FFNO} is adapted from FNO and contains an improved representation layer for the operator and a better set of training approaches. Factorization is adopted in the Fourier layer to reduce the number of parameters. Our implementation of FFNO is adapted from the official implementation\footnote{https://github.com/alasdairtran/fourierflow}.
    \item CNO~\cite{CNO} is proposed as a modification of convolutional neural network to enable effective operator learning. It is instantiated as a novel operator adaptation of U-Net~\cite{U-Net}. Our implementation of CNO is adapted from the official implementation\footnote{https://github.com/bogdanraonic3/ConvolutionalNeuralOperator}.
\end{itemize}

% Appendix \ref{sec:appendix_hyperparam} provides detailed hyperparameters for these baselines. In SpecB-FNO, the multiple FNOs use the same hyperparameters. 

\subsection{Hyperparameter Settings}
\label{sec:appendix_hyperparam}

This section focuses on the hyperparameters used in our experiments for FNOs, including layers, frequency modes, hidden channels, learning rate, etc. We report the hyperparameters adopted in Table \ref{tab:main} in Table \ref{tab:hyper}. 
% Note that for SpecB-FNO, its hyperparameters are the same as FNO without explicit specification.

\begin{table}[!htbp]
\centering
\caption{Hyperparameter in Table \ref{tab:main}}
\label{tab:hyper}
% \resizebox{\textwidth}{!}{
\begin{tabular}{c|c|c|c|c|c|c}
\hline
    % \multirow{2}{*}{Model} & \multirow{2}{*}{Hyperparam} & \multirow{2}{*}{Darcy Flow} & \multicolumn{2}{c|}{Navier-Stokes} & \multirow{2}{*}{shallow water} & \multirow{2}{*}{Diffusion Reaction} \\
    \multirow{2}{*}{Model} & Hyper- & Darcy & \multicolumn{2}{c|}{Navier-Stokes} & shallow & diffusion- \\
\cline{4-5}
    % & & & $\nu$ = 1e-3 & $\nu$ = 1e-5 & & \\
    & Parameter & flow & $\nu$ = 1e-3 & $\nu$ = 1e-5 & water & reaction \\
\hline
    General & batch size & 20 & 40 & 40 & 40 & 40 \\
\hline
    \multirow{4}{*}{DeepONet} 
    & layer & 3 & 3 & 3 & 3 & - \\
    & channel & 160 & 200 & 200 & 160 & - \\
    & lr & 1e-3 & 1e-3 & 1e-3 & 1e-3 & - \\
    & epoch & 50 & 35 & 100 & 20 & - \\
\hline
    \multirow{4}{*}{ResNet}
    & layer & 18 & 18 & 18 & 10 & 10 \\
    & channel & 30 & 30 & 50 & 30 & 30 \\
    & lr & 1e-3 & 1e-3 & 1e-3 & 1e-3 & 1e-3 \\
    & epoch & 30 & 20 & 70 & 15 & 56 \\
\hline
    \multirow{4}{*}{U-Net}
    & layer & 8 & 8 & 8 & 8 & 8 \\
    & channel & 20 & 40 & 60 & 30 & 20 \\
    & lr & 1e-4 & 2e-4 & 2e-4 & 1e-4 & 1e-4 \\
    & epoch & 40 & 35 & 100 & 15 & 64 \\
\hline
    \multirow{4}{*}{CNO}
    & layer & 8 & 8 & 8 & 8 & 8 \\
    & channel & 60 & 60 & 90 & 30 & 30 \\
    & lr & 1e-3 & 1e-3 & 1e-3 & 1e-3 & 1e-3 \\
    & epoch & 30 & 25 & 70 & 20 & 64 \\
% \hline
%     \multirow{5}{*}{FNO \& FFNO}
%     & layer & 8 & 8 & 8 & 6 & 6 \\
%     & channel & 60 & 60 & 100 & 40 & 30 \\
%     & lr & 2e-4 & 1e-4 & 2e-4 & 1e-4 & 1e-4 \\
%     & epoch & 30 & 35 & 100 & 40 & 80 \\
%     & modes & 24 & 16 & 32 & 24 & 64 \\
\hline
    \multirow{5}{*}{FNO} 
    & layer & 8 & 8 & 8 & 6 & 6 \\
    & channel & 60 & 60 & 100 & 40 & 30 \\
    & lr & 2e-4 & 1e-4 & 2e-4 & 1e-4 & 1e-4 \\
    & epoch & 30 & 35 & 100 & 40 & 80 \\
    & modes & 8 & 8 & 16 & 24 & 32 \\
\hline
    \multirow{5}{*}{FFNO} 
    & layer & 8 & 8 & 8 & 6 & 6 \\
    & channel & 60 & 60 & 100 & 40 & 30 \\
    & lr & 2e-4 & 1e-4 & 2e-4 & 1e-4 & 1e-4 \\
    & epoch & 30 & 35 & 100 & 40 & 80 \\
    & modes & 24 & 16 & 32 & 64 & 64 \\
\hline
    \multirow{6}{*}{SpecB-FNO}
    & layer & 8 & 8 & 8 & 6 & 6 \\
    & channel & 60 & 60 & 100 & 40 & 30 \\
    & lr & 2e-4 & 1e-4 & 2e-4 & 1e-4 & 1e-4 \\
    & epoch & 30 & 35 & 100 & 40 & 80 \\
    & modes & 64 & 16 & 32 & 64 & 64 \\
    & T & 3 & 1 & 1 & 1 & 1 \\
\hline
\end{tabular}
% }
\end{table}

\subsection{Hardware and Computing}
\label{sec:appendix_hardware}
All experiments are conducted on a DGX server with 40 Intel(R) Xeon(R) CPU E5-2698 v4 2.20GHz CPUs, 4 Tesla V100-DGXS-32GB GPUs, and 251 GB memory. 

% Each experiment in this paper uses only one GPU and up to four CPUs. 
The memory consumption for each experiment is less than 50GB, and the time required to train each surrogate model is no more than three hours. 

\subsection{Code Implementation}
\label{sec:appendix_code}

Our codebase is contained in the supplementary material. 

% {\color{red} We append the skeleton code repository in the appended file. Our model configurations are provided in previous sections. We will open-source the repository and ensure the reproducibility of our method upon acceptance.}

\section{Ablation Study on SpecB-FNO over Different FNO Configurations}

In this section, we conduct an ablation study on the effectiveness of SpecB-FNO over different configurations. Such an experiment can be utilized to (i) illustrate the optimal of our hyper-parameter selection and (ii) the empirical observation in Section \ref{sec:spectral} that enlarging the Fourier kernel for FNO does not necessarily lead to better accuracy.

We config FNO with two key hyperparameters, namely layers and frequency modes, on Navier-Stokes datasets with $\nu$ = 1e-3 and $\nu$ = 1e-5. The results are shown in Table \ref{tab:search-3} and Table \ref{tab:search-5}, respectively.

% We conduct a grid search on two key hyperparameters of FNO, namely layers and frequency modes, on Navier-Stokes datasets with $\nu$ = 1e-3 and $\nu$ = 1e-5. The purpose of showing the grid search result is to demonstrate 

\begin{table*}[!htbp]
% \tiny
\centering
\caption{Relative error (\%) comparison on Navier-Stokes ($\nu$ = 1e-3) between FNO and SpecB-FNO utilizing FNO-skip with different layers and frequency modes. The hidden channels of FNO-skip are set to 60. Imp. indicates the relative improvement from FNO to SpecB-FNO.}
\vspace{0.5\baselineskip}
% \begin{tabular*}{\linewidth}{@{\extracolsep{\fill}} l | *{3}{c} | *{3}{c} | *{3}{c}}
\resizebox{\textwidth}{!}{
\begin{tabular}{l | *{3}{c} | *{3}{c} | *{3}{c}}  
\toprule[+1pt]
\multirow{2}{*}{Layer} & \multicolumn{3}{c|}{modes = 8} & \multicolumn{3}{c|}{modes = 16} & \multicolumn{3}{c}{modes = 32} \\
\cmidrule(lr){2-4} \cmidrule(lr){5-7} \cmidrule(lr){8-10}
& FNO & SpecB-FNO &  Imp. (\%) & FNO & SpecB-FNO &  Imp. (\%) & FNO & SpecB-FNO &  Imp. (\%) \\
\midrule 
4    & 0.47 $\pm$ 0.03 & 0.20 $\pm$ 0.01 & 57.1 & 1.73 $\pm$ 0.40 & 0.32 $\pm$ 0.07 & 81.3 & 1.91 $\pm$ 0.31 & 0.30 $\pm$ 0.02 & 84.1\\
8    & 0.39 $\pm$ 0.03 & 0.17 $\pm$ 0.01 & 55.1 & 0.47 $\pm$ 0.01 & 0.14 $\pm$ 0.01 & 69.7 & 0.45 $\pm$ 0.01 & 0.18 $\pm$ 0.02 & 59.5\\
16   & 0.40 $\pm$ 0.01 & 0.20 $\pm$ 0.01 & 49.1 & 0.46 $\pm$ 0.01 & 0.21 $\pm$ 0.02 & 54.8 & 0.41 $\pm$ 0.01 & 0.19 $\pm$ 0.02 & 52.5\\
\bottomrule[+1pt]
\end{tabular}
}
% \end{tabular*}
\label{tab:search-3}
\end{table*}

Table \ref{tab:search-3} illustrates the impact of frequency modes and layers on Navier-Stokes with $\nu$ = 1e-3. Increasing frequency modes fails to enhance FNO's performance. SpecB-FNO can perform better with a larger frequency mode of 16. Keep increasing the frequency mode for SpecB-FNO doesn't bring more improvements, because Navier-Stokes with $\nu$ = 1e-3 contains negligible high-frequency information. While increasing layers from 4 to 8 yields performance improvements for both FNO and SpecB-FNO, further increments to 16 don't provide additional benefits, likely due to the risk of overfitting with deeper models.

\begin{table*}[!htbp]
% \tiny
\centering
\caption{Relative error (\%) comparison on Navier-Stokes ($\nu$ = 1e-5) between FNO and SpecB-FNO utilizing FNO-skip with different layers and frequency modes. The hidden channels of FNO-skip are set to 100. Imp. indicates the relative improvement from FNO to SpecB-FNO.}
\vspace{0.5\baselineskip}
% \begin{tabular*}{\linewidth}{@{\extracolsep{\fill}} l | *{3}{c} | *{3}{c} | *{3}{c}}
\resizebox{\textwidth}{!}{
\begin{tabular}{l | *{3}{c} | *{3}{c} | *{3}{c}}  
\toprule[+1pt]
\multirow{2}{*}{Layer} & \multicolumn{3}{c|}{modes = 8} & \multicolumn{3}{c|}{modes = 16} & \multicolumn{3}{c}{modes = 32} \\
\cmidrule(lr){2-4} \cmidrule(lr){5-7} \cmidrule(lr){8-10}
& FNO & SpecB-FNO &  Imp. (\%) & FNO & SpecB-FNO &  Imp. (\%) & FNO & SpecB-FNO &  Imp. (\%) \\
\midrule 
4       & 6.64 $\pm$ 0.03 & 6.21 $\pm$ 0.06 & 6.45 & 6.55 $\pm$ 0.10 & 4.84 $\pm$ 0.03 & 26.0 & 6.97 $\pm$ 0.06 & 5.60 $\pm$ 0.04 & 19.6\\
8       & 6.07 $\pm$ 0.03 & 5.73 $\pm$ 0.05 & 5.63 & 5.76 $\pm$ 0.04 & 3.92 $\pm$ 0.03 & 31.9 & 6.03 $\pm$ 0.05 & 3.51 $\pm$ 0.14 & 41.8\\
16      & 6.18 $\pm$ 0.07 & 6.01 $\pm$ 0.41 & 2.71 & 5.82 $\pm$ 0.13 & 4.08 $\pm$ 0.12 & 29.9 & 5.94 $\pm$ 0.05 & 3.64 $\pm$ 0.64 & 38.7\\
\bottomrule[+1pt]
\end{tabular}
}
% \end{tabular*}
\label{tab:search-5}
\end{table*}

Table \ref{tab:search-5} illustrates the impact of frequency modes and layers on Navier-Stokes with $\nu$ = 1e-5. Notably, increasing frequency modes improves SpecB-FNO's performance, whereas FNO remains unaffected. This disparity arises from SpecB-FNO's ability to leverage higher frequency modes in Fourier layers, a benefit not accessible to FNO due to its Fourier parameterization bias. Similarly to Table \ref{tab:search-3}, while increasing layers from 4 to 8 yields performance improvements for both FNO and SpecB-FNO, further increments to 16 don't provide additional benefits.

\section{One-step Error for Solving PDE}
\label{sec:appendix-effective-one}

For sequential PDE datasets, Table \ref{tab:main} presents the average error across the entire prediction sequence. To facilitate a better understanding of SpecB-FNO, we report the average one-step prediction error in Table \ref{tab:main-one}. Compared to Table \ref{tab:main}, the performance gaps between different surrogate models in Table \ref{tab:main-one} are smaller. This is because auto-regressive prediction can lead to accumulative errors, which is further discussed in Appendix ~\ref{sec:appendix_cumulate}. Additionally, we can observe that even if two surrogate models perform similarly at the first step, their performance gap can become large after many steps of auto-regressive prediction. Hence, we argue that cumulative error can better evaluate a model than one-step error, similar to previous works~\cite{FNO,UNO,MP-PDE}.

\begin{table}[htbp]
\centering
\vspace{-5pt}
\caption{One-step Error Comparison between SpecB-FNO and Baselines}   \label{tab:main-one}
\begin{tabular}{c|cccc}
\hline
    \multirow{2}{*}{Model} & \multicolumn{2}{c}{Navier-Stokes} & shallow & diffusion- \\
\cline{2-3}
    & $\nu$ = 1e-3 & $\nu$ = 1e-5 & water & reaction \\
\hline
    DeepONet  & .0335$\pm$.0008 & .2176$\pm$.0055 & .1364$\pm$.0176 & NaN \\
\hline
    ResNet    & .0133$\pm$.0002 & .1157$\pm$.0019 & .0264$\pm$.0024 & .0042$\pm$.0001 \\
    U-Net     & .0056$\pm$.0001 & .0682$\pm$.0008 & .0296$\pm$.0034 & .0359$\pm$.0006 \\
    CNO       & .0057$\pm$.0001 & .0528$\pm$.0007 & .0111$\pm$.0012 & .0033$\pm$.0036 \\
\hline
    FNO       & .0008$\pm$.0001 & .0355$\pm$.0004 & .0029$\pm$.0000 & .0042$\pm$.0001 \\
    FFNO      & .0065$\pm$.0001 & .0505$\pm$.0007 & .0189$\pm$.0009 & .0023$\pm$.0001 \\
    SpecB-FNO & .0003$\pm$.0000 & .0242$\pm$.0016 & .0002$\pm$.0000 & .0013$\pm$.0001 \\
\hline
\end{tabular}
% }
\begin{tablenotes}
\footnotesize
\item[1] \textit{NaN} indicates that the experiment does not converge.
\end{tablenotes}
\vspace{-5pt}
\end{table}

\section{PDE Data Reconstruction Experiments}
% \section{FNO-based Superresolution (FNO-SR) Model and FNO-based Autoencoder (FNO-AE)}
\label{sec:appendix_sr}

In this section, we evaluate the performance of SpecB-FNO on a different task, Data Reconstruction, over PDE data. We first introduce the two variants for data reconstruction in Section \ref{sec:appendix_sr_model}. Then we report the empirical result in Section \ref{sec:appendix_sr_experiment}

\subsection{FNO-based Superresolution (FNO-SR) Model and FNO-based Autoencoder (FNO-AE)}
\label{sec:appendix_sr_model}

\begin{figure}[!htbp]
    \centering
    \includegraphics[width=0.7\columnwidth]{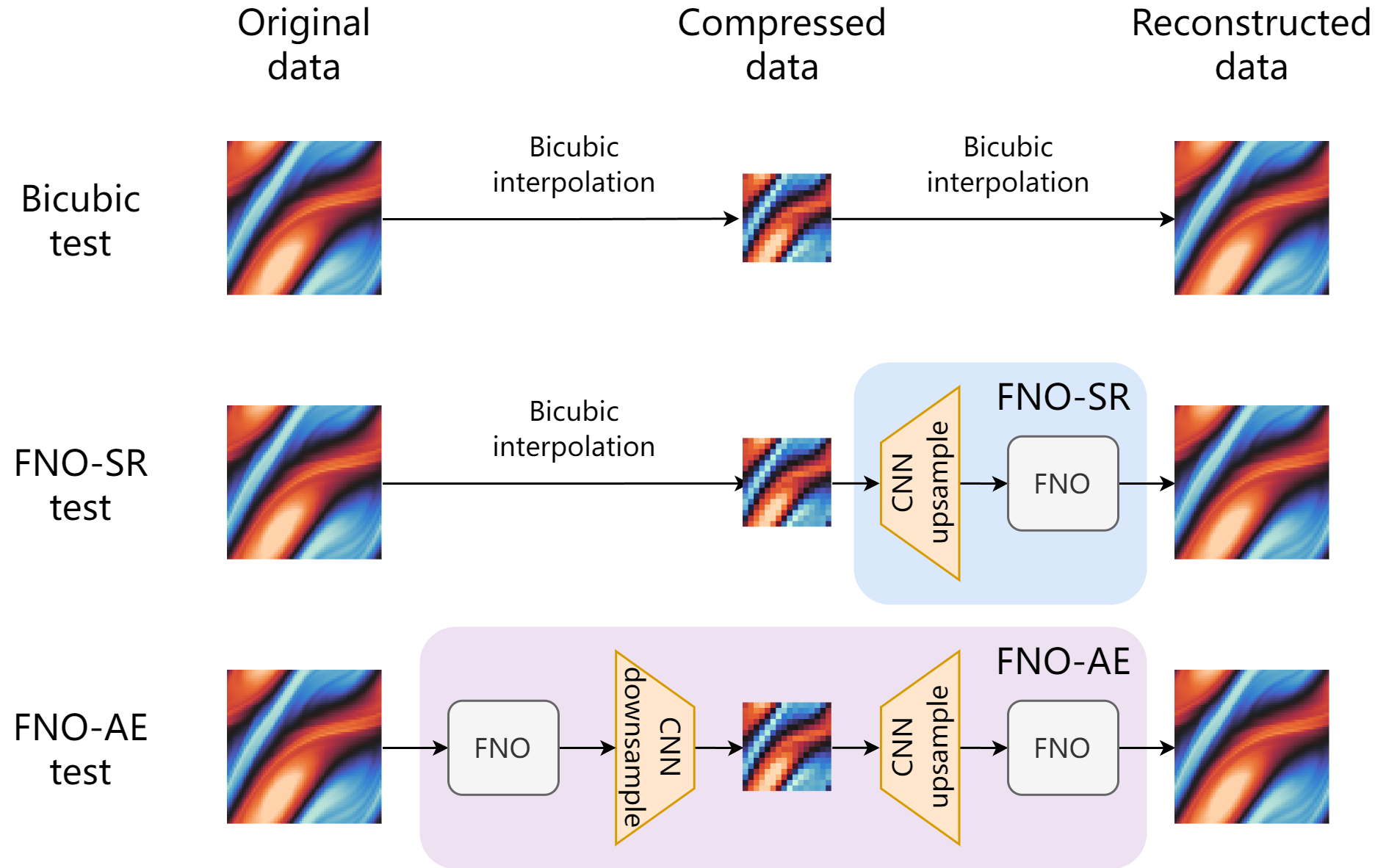}
    \vspace{-5pt}
    \caption{Illustration on data reconstruction experiment and architectures of FNO-SR and FNO-AE.}
    \vspace{1pt}
    \label{fig:sr}
\end{figure}

While FNO is crafted to be a resolution-invariant model, it always requires identical resolution for its input and output. As a result, FNO cannot take a low-resolution input to predict a high-resolution output or vice versa. To enable upsampling and downsampling for FNO, we integrate convolution layers into both the FNO-based superresolution model (FNO-SR) and the FNO-based autoencoder (FNO-AE), as illustrated in Figure \ref{fig:sr}. 

FNO-SR and FNO-AE adopt a straightforward design, incorporating basic CNN layers for downsampling or upsampling. These layers are placed before the initial or after the final FNO layer, maintaining FNO's internal architecture. The FNO-skip block is employed for both FNO-SR and FNO-AE.

In the experiment in Table \ref{tab:nslayer}, FNO means directly training a solo FNO model. SpecB-FNO means sequentially training two models. In the case of SpecB-FNO applied to FNO-AE, two FNO-AEs generate two sets of latent variables, resulting in doubling the latent variable size. To ensure a fair comparison, the SpecB-FNO at a compression ratio of 2:1 is an ensemble of two FNO-AEs with a compression ratio of 4:1, for example. 

\subsection{PDE Data Reconstruction}
\label{sec:appendix_sr_experiment}

In addition to solving PDEs, we further explore SpecB-FNO's effectiveness for PDE data compression and reconstruction. 
Compressing \cite{rowley2017model} and reconstructing \cite{fukami2019super} PDE simulation data are pivotal in advancing fluid dynamics research.
% The reconstruction of PDE simulation data from low to high resolution plays a pivotal role in fluid dynamics research \cite{fukami2019super}. 
% Our evaluation aims to highlight SpecB-FNO's effectiveness on FNO in reconstructing PDE data rather than pursuing the perfect model design for PDE data reconstruction.
We assess the compression and reconstruction capabilities of SpecB-FNO on the 2D Navier-Stokes dataset with $\nu$ = 1e-5. The evaluation involves compressing the flow field to a lower resolution and reconstructing it to the original resolution, aiming to minimize the reconstruction error. We compare the following three methods:
(i) \textbf{Bicubic}: compression and reconstruction of data using bicubic interpolation.
(ii) \textbf{FNO-SR}: compression of data with bicubic interpolation, followed by reconstruction using an FNO-based superresolution model.
(iii) \textbf{FNO-AE}: compression and reconstruction of data using an FNO-based autoencoder.
Convolutional layers are additionally stacked with the input layer or the output layer of the FNO to enable upsampling or downsampling. Details of the model architecture are provided in Appendix \ref{sec:appendix_sr}.

\begin{table}[!htbp]
\caption{Relative error (\%) comparison on Navier-Stokes ($\nu$=1e-5) data reconstruction between FNO and SpecB-FNO with FNO-SR and FNO-AE. FNO  indicates using the standard single FNO for SR and AE. SpecB-FNO indicates sequentially training two FNOs for SR and AE. Imp. indicates the relative improvement from FNO to SpecB-FNO. CR. indicates the data compression ratio.}
% \vspace{0.5\baselineskip}
\centering
% \resizebox{\columnwidth}{!}{
\begin{tabular}{l|cccc}
\toprule [+1pt]
CR. & Method & Bicubic & FNO-SR & FNO-AE \\
\midrule 
\multirow{3}{*}{2 $:$ 1} 
& FNO       & 2.70  & 4.98 $\pm$ 0.08       & 1.89 $\pm$ 0.06 \\
& SpecB-FNO & -     & 4.28 $\pm$ 0.08       & \bf{1.14 $\pm$ 0.03} \\
& Imp. (\%) & -     & 14.1                  & \bf{39.7} \\
\cmidrule(lr){1-5}
\multirow{3}{*}{4 $:$ 1} 
& FNO       & 4.78  & 2.70 $\pm$ 0.01       & 2.51 $\pm$ 0.05 \\
& SpecB-FNO & -     & \bf{1.56 $\pm$ 0.01}  & 1.82 $\pm$ 0.07 \\
& Imp. (\%) & -     & \bf{42.2}             & 27.5 \\
\cmidrule(lr){1-5}
\multirow{3}{*}{8 $:$ 1} 
& FNO       & 7.51  & 3.90 $\pm$ 0.01       & 3.32 $\pm$ 0.09 \\
& SpecB-FNO & -     & 2.98 $\pm$ 0.03       & \bf{2.70 $\pm$ 0.04} \\
& Imp. (\%) & -     & \bf{23.6}             & 18.7 \\
\cmidrule(lr){1-5}
\multirow{3}{*}{16 $:$ 1} 
& FNO       & 11.54 & 5.21 $\pm$ 0.04       & 4.36 $\pm$ 0.08 \\
& SpecB-FNO & -     & 4.82 $\pm$ 0.03       & \bf{4.35 $\pm$ 0.08} \\
& Imp. (\%) & -     & \bf{7.5}              & 0.2 \\
\bottomrule [+1pt]
\end{tabular}
% }
% \vspace{-10pt}
\label{tab:sr_transposed}
\end{table}

Table \ref{tab:sr_transposed} reports the performance of different configurations on the N-S dataset. We can make the following three observations:
To begin, SpecB-FNO consistently outperforms FNO in all scenarios, aligning with our findings from previous sections. Secondly, FNO-AE exhibits superior performance compared to FNO-SR. The ability of FNO-AE to learn a more effective representation surpasses Bicubic, which is the downsampling component when testing FNO-SR. Third, As the compression ratio increases, more information is lost during the compression. Hence, the performance of FNO and SpecB-FNO both decreases. Finally, as the compression ratio increases, the relative improvements of SpecB-FNO compared to FNO decrease, as high-frequency information is more likely to be discarded during compression. With less high-frequency information, the superiority of SpecB-FNO against FNO is less evident.

\section{Error accumulation on SpecB-FNO}
\label{sec:appendix_cumulate}

Since we employ autoregressive prediction with one-step input and one-step output on Navier-Stokes, the prediction error accumulates as the sequential index $t$ increases. We report the averaged result and visualization of error accumulation in the experiment of Table \ref{tab:main} for FNO and SpecB-FNO in Section \ref{sec:appendix_cumulate_result} and \ref{sec:appendix_cumulate_visual}, respectively.

\subsection{Result of Error Accumulation}
\label{sec:appendix_cumulate_result}

\begin{figure}
\centering
\begin{subfigure}{0.45\textwidth}
    \includegraphics[width=\textwidth]{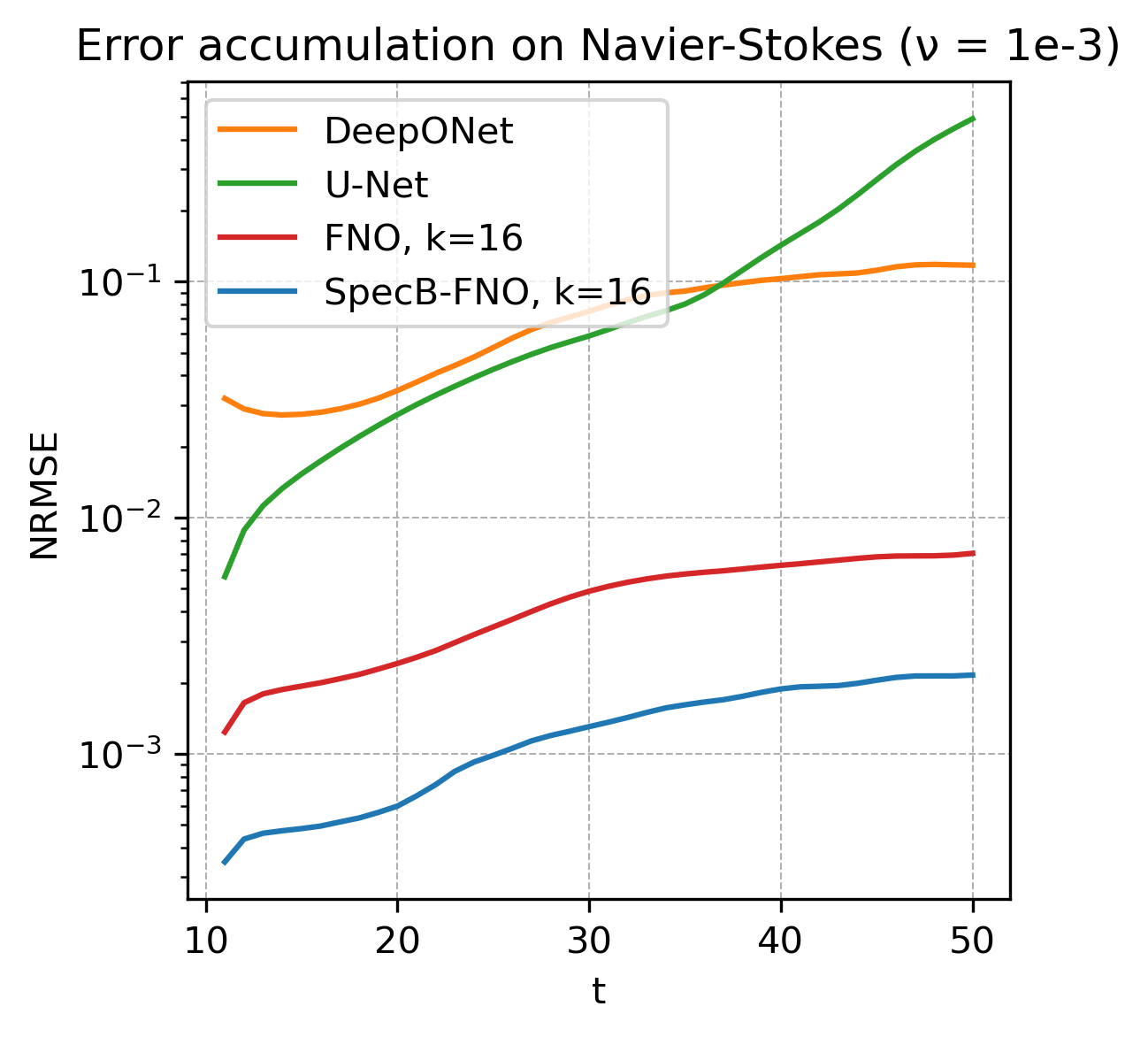}
    \caption{$\nu$ = 1e-3}
    \label{fig:curve-t-3}
\end{subfigure}
\hfill
\begin{subfigure}{0.45\textwidth}
    \includegraphics[width=\textwidth]{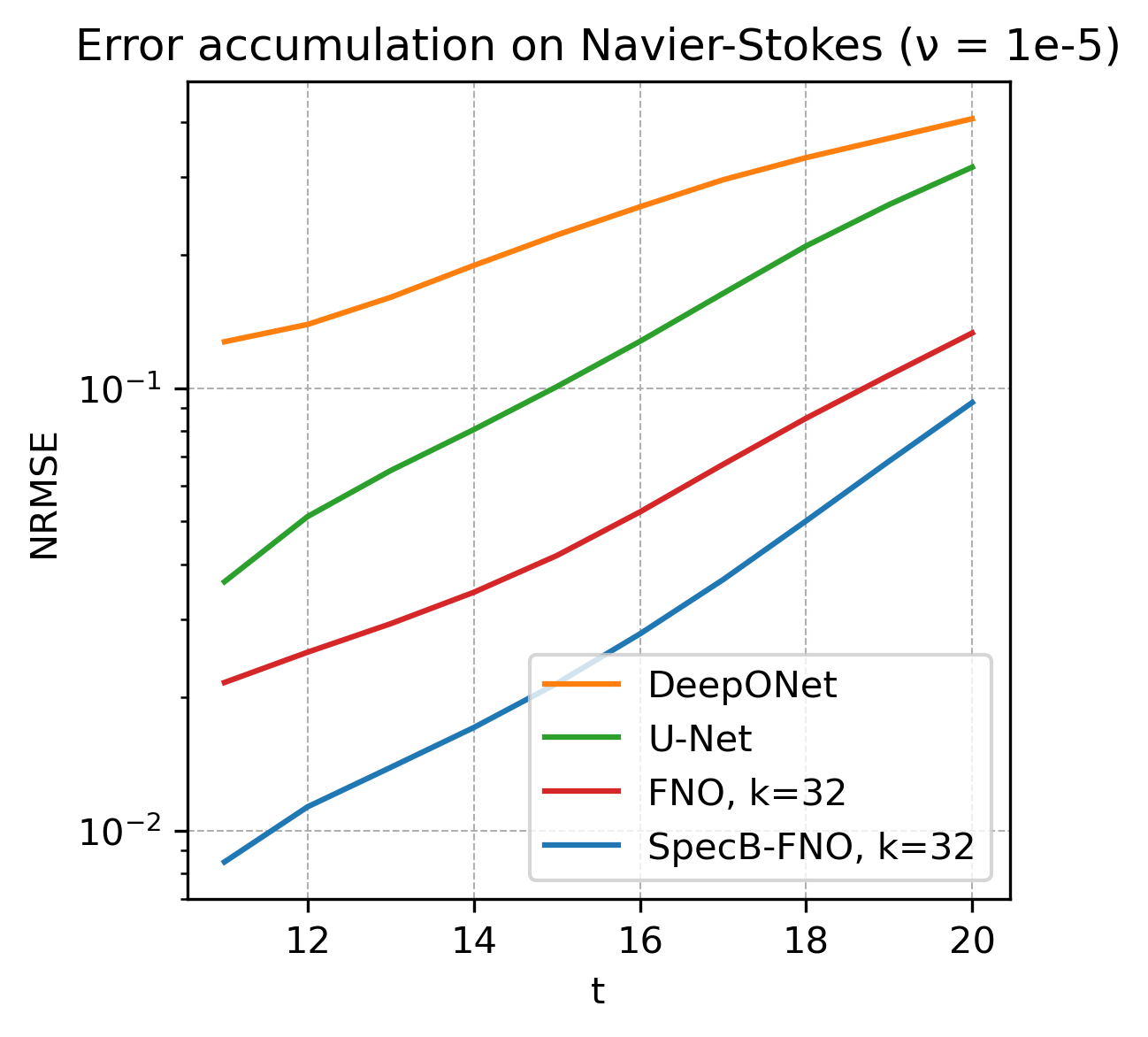}
    \caption{$\nu$ = 1e-5}
    \label{fig:curve-t-5}
\end{subfigure}
\caption{Relative error (\%) accumulation comparison on Navier-Stokes. t denotes the sequential index in the Navier-Stokes dataset.}
\label{fig:accumulative}
\end{figure}

We first report the average error at different steps $t$ over the test set. The result of the Navier-Stokes equation with $\nu$ equals 1e-3 and 1e-5 are illustrated in Figures \ref{fig:curve-t-5} and \ref{fig:curve-t-3}. We can easily make the following observations.

First, as the step $t$ increases, both FNO and SpecB-FNO accumulate prediction errors. 
Second, SpecB-FNO constantly outperforms FNO, indicating its effectiveness during long-term prediction.
Third, due to the distinct spectral behaviors of SpecB-FNO with $\nu$ equals 1e-3 and 1e-5, its influence on error accumulation differs. On the dataset with $\nu$ = 1e-5, the enhancement provided by SpecB-FNO tends to diminish as error accumulates. This phenomenon is attributed to the fact that long-term prediction error is more closely tied to the low-frequency components in the data \cite{davidson2015turbulence}, and SpecB-FNO's improvement in low-frequency accuracy is limited when $\nu$ = 1e-5 (Figure \ref{fig:spectral-ns-5}). Conversely, when $\nu$ = 1e-3, SpecB-FNO reduces both low-frequency and high-frequency residuals (Figure \ref{fig:spectral-ns-3}), resulting in an improvement conducive to long-term prediction.

\subsection{Visualization for Error Accumulation}
\label{sec:appendix_cumulate_visual}

\begin{figure}[!htbp]
    \centering
    \includegraphics[width=0.98\textwidth]{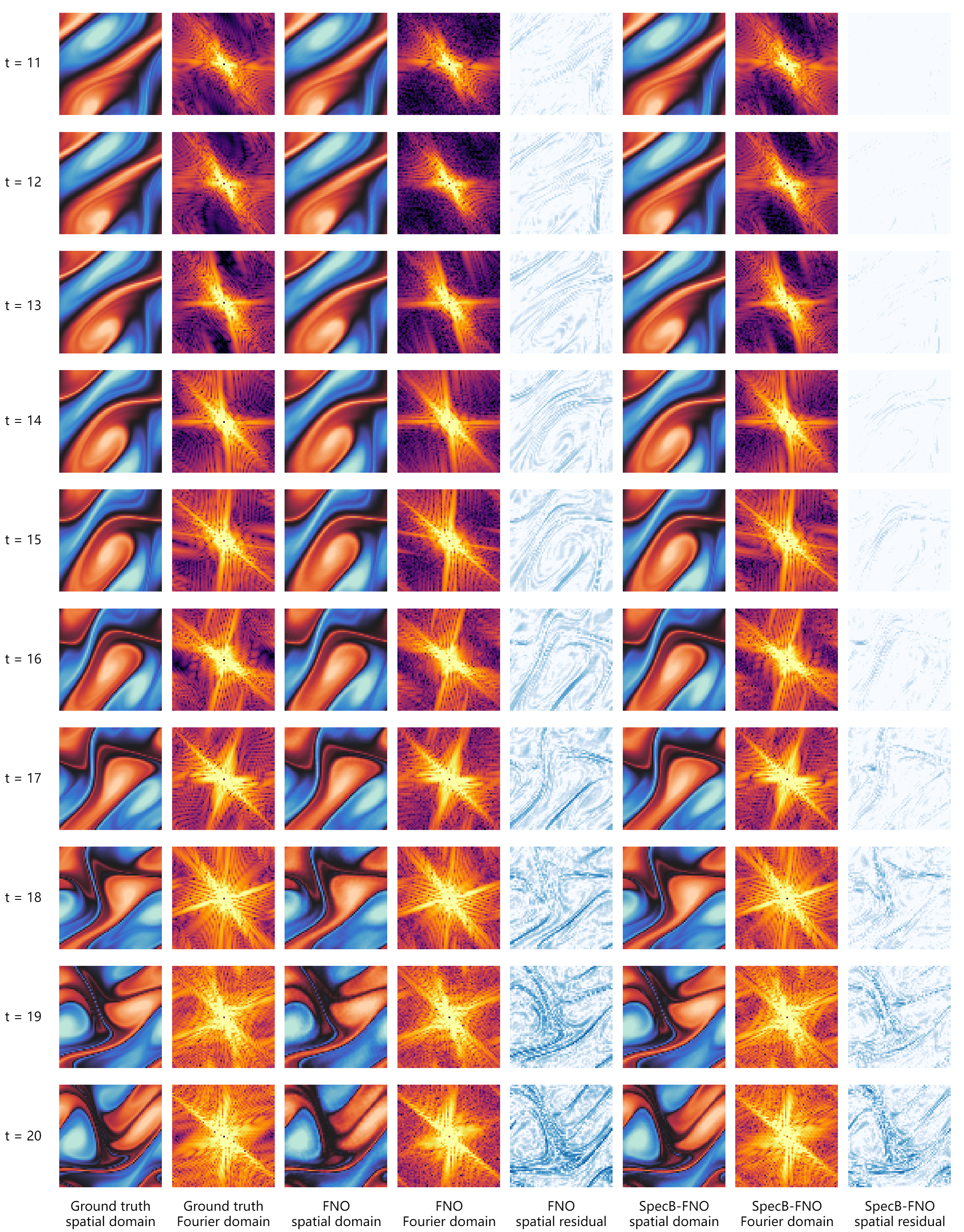}
    \vspace{-5pt}
    \caption{Visualization for error accumulation on Navier-Stokes ($\nu$ = 1e-5) with FNO, k=32 and SpecB-FNO, k=32.}
    \label{fig:long-5}
\end{figure}

\begin{figure}[!htbp]
    \centering
    \includegraphics[width=0.98\textwidth]{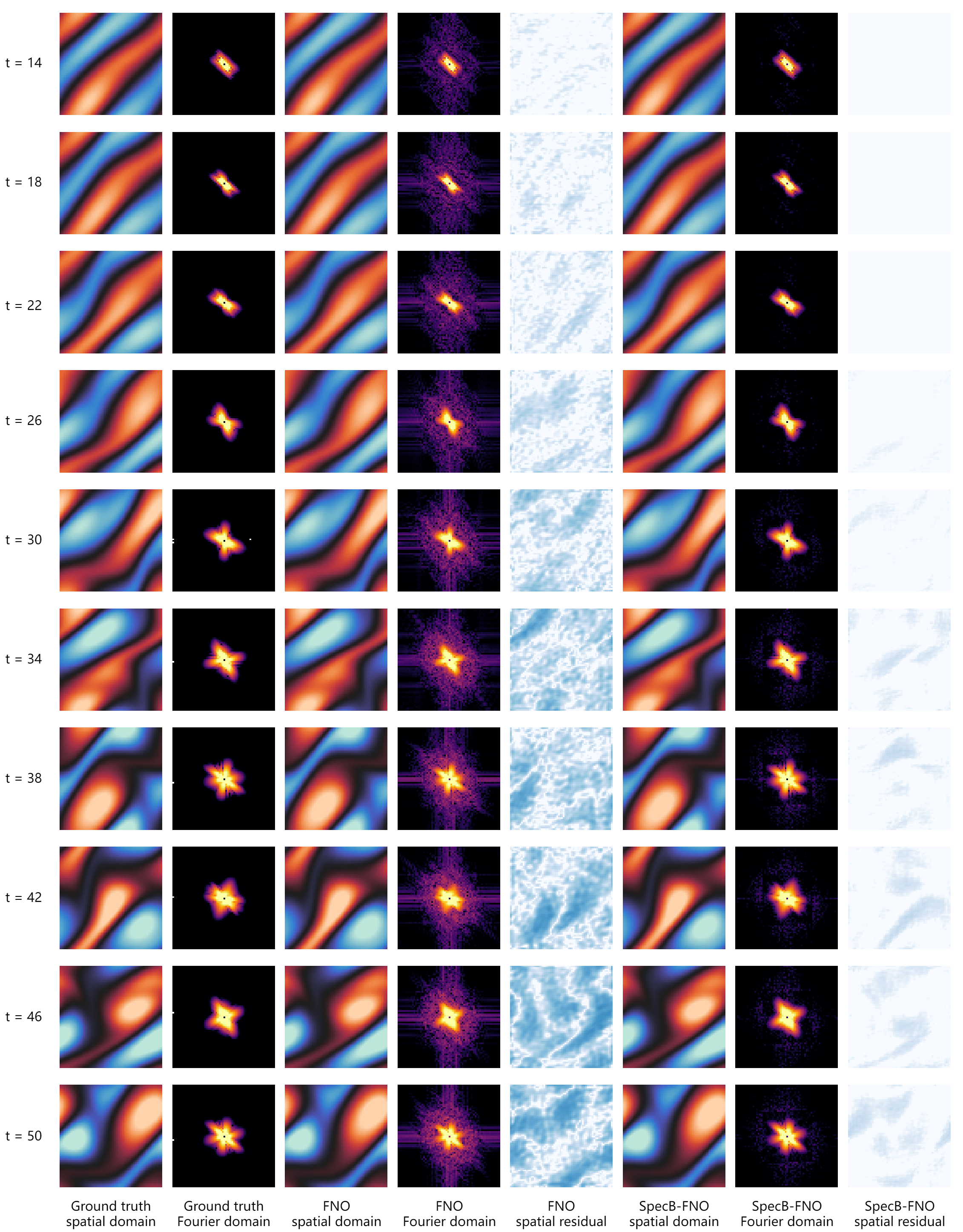}
    \vspace{-5pt}
    \caption{Visualization for error accumulation on Navier-Stokes ($\nu$ = 1e-3) with FNO, k=16 and SpecB-FNO, k=16.}
    \label{fig:long-3}
\end{figure}

Here, we present the visualization of error accumulation on both spatial and spectral domains in the experiment of Figures \ref{fig:spectral-ns-5} and \ref{fig:spectral-ns-3} for FNO and SpecB-FNO. The results for $\nu$ value at 1e-5 and 1e-3 are shown in Figure \ref{fig:long-5} and \ref{fig:long-3}, respectively. % We can easily come to the same conclusion. 
% Since we employ autoregressive prediction with one-step input and one-step output on Navier-Stokes, the prediction error accumulates as the sequential index $t$ increases. We present the visualization of error accumulation in the experiment of Table \ref{tab:main} for  and SpecB-FNO with FNO-skip. The results for $\nu$ value at 1e-5 and 1e-3 are shown in Figure \ref{fig:long-5} and \ref{fig:long-3}, respectively. We can observe that as the step $t$ increases, both  and SpecB-FNO accumulate prediction errors. Besides, SpecB-FNO constantly outperforms , indicating its effectiveness during long-term prediction.

In Figure \ref{fig:long-3}, the PDE data resolution is 64 $\times$ 64. Both FNO and SpecB-FNO have a truncation frequency of 16, resulting in Fourier kernels of size 32 $\times$ 32. For FNO, high-frequency noise within the Fourier kernel is clearly visible in the spectral domain, illustrating FNO's ineffectiveness due to its Fourier parameterization bias. In contrast, SpecB-FNO significantly reduces the high-frequency noise within the Fourier kernel.

%%%%%%%%%%%%%%%%%%%%%%%%%%%%%%%%%%%%%%%%%%%%%%%%%%%%%%%%%%%%

\end{document}